\documentclass{article}
\usepackage{spconf,amsmath,graphicx}
\usepackage{xcolor}
\usepackage{amssymb}
\usepackage{float}
\usepackage{placeins}
\usepackage{booktabs}
\usepackage{multirow}
\usepackage{soul}
\usepackage{subcaption}

\definecolor{customgreen}{RGB}{0,170,85}
\newcommand{\best}[1]{\textbf{\textcolor{blue}{#1}}}

\title{AquaDiff: Diffusion-Based Underwater Image Enhancement for Addressing Color Distortion}

\name{Afrah Shaahid$^{1}$ \quad Muzammil Behzad$^{1,2,*}\thanks{*Corresponding author. Email: \texttt{muzammil.behzad@kfupm.edu.sa}}$}

\address{
$^{1}$King Fahd University of Petroleum and Minerals, Saudi Arabia\\
$^{2}$SDAIA--KFUPM Joint Research Center for Artificial Intelligence, Saudi Arabia
}

\begin{document}
\maketitle

\begin{abstract}

Underwater images are severely degraded by wavelength-dependent light absorption and scattering, resulting in color distortion, low contrast, and loss of fine details that hinder vision-based underwater applications. To address these challenges, we propose AquaDiff, a diffusion-based underwater image enhancement framework designed to correct chromatic distortions while preserving structural and perceptual fidelity. AquaDiff integrates a chromatic prior-guided color compensation strategy with a conditional diffusion process, where cross-attention dynamically fuses degraded inputs and noisy latent states at each denoising step. An enhanced denoising backbone with residual dense blocks and multi-resolution attention captures both global color context and local details. Furthermore, a novel cross-domain consistency loss jointly enforces pixel-level accuracy, perceptual similarity, structural integrity, and frequency-domain fidelity. Extensive experiments on multiple challenging underwater benchmarks demonstrate that AquaDiff provides good results as compared to the state-of-the-art traditional, CNN-, GAN-, and diffusion-based methods, achieving superior color correction and competitive overall image quality across diverse underwater conditions.
\end{abstract}

\begin{keywords}
Artificial Intelligence, Computer Vision, Deep Learning, Underwater Image Enhancement, Diffusion Models
\end{keywords}

\section{Introduction}
\label{sec:intro}

Underwater imagery often exhibits degraded quality marked by intense color shifts, muted contrast, and softened details. These artifacts stem from attenuation and scattering effects that vary with wavelength and distance as light travels through water, ultimately impairing the performance of vision-based underwater systems. 

In clear water, light is absorbed and scattered so strongly that only blue-green wavelengths penetrate far, causing a blue tint and low contrast \cite{saoud2024seeing, qi2021underwater, sun2023high}. In practice, water absorbs long that is red wavelengths first and only transmits shorter that are green/blue or light, so reds vanish quickly as depth increases. Meanwhile both forward- and back-scattering by water itself and dissolved matter blurs the scene. The combined effect of this wavelength-dependent absorption and volume scattering is dim, desaturated images with a foggy veiling layer and blurred features \cite{hao2024underwater}.

Biological and environmental factors add further haze. Tiny organisms and debris known as marine snow and clouds of sediment act like moving specular flakes in the water. High turbidity such as plankton blooms, stirred sediment, or organic detritus creates a white veiling scatter that overwhelms faint scene radiance \cite{galetto2025deep, alsakar2025underwater, zhou2022yolotrashcan, cao2022dynamic}. These particulates multiply the scattering and absorption. The backscattered light from all this matter reduces contrast and dims distant objects, making the entire scene appear flat and murky. In very turbid water, nothing beyond a short range is visible and useful image features are lost.

Hardware and sensor constraints also limit image quality. Typical underwater cameras have only moderate dynamic range and must operate in low light, so shadows often fall into sensor noise while bright areas saturate. Moreover the optics and housings introduce distortions so most underwater cameras use a flat glass port or dome which refracts the incoming light before it hits the lens \cite{naveen2025advancements,almutiry2024underwater}.

The degraded underwater images effects a broad range of tasks including object detection and classification algorithms. This leads to miss or mis-labeling of marine organisms \cite{zhou2025uw}. Moreover, SLAM and navigation pipelines lose track in murky video, and 3D photogrammetric reconstructions \cite{huang2025visual} suffer from sparse points and geometric errors. Using the degraded underwater images, the underwater robots and monitoring systems struggle to detect, track, navigate and map underwater scenes accurately.

To counter these limitations, the field of underwater image enhancement (UIE) focuses on recovering lost detail and improving the perceptual fidelity of underwater scenes, and has been extensively explored over the past several decades. Recent studies integrate specialized image enhancement steps to recover visual fidelity. By correcting color casts and increasing contrast, enhancement algorithms restore the subtle texture and color cues that detectors and feature matchers need \cite{summers2025impact}. For example, a study \cite{zhang2023improved} stated that that adding a color-correction preprocessing to a YOLOv5-based marine object detector alleviates the interference of underwater image degradation. This further yielded approximately a 3–4\% absolute improvement in mAP on standard coral-monitoring datasets. 

Another study \cite{summers2025impact} demonstrated that enhanced frames lead to much more robust SLAM. The restored underwater images produced more consistent keypoint matches and more accurate loop closures in underwater trajectories. One of the recent literature reviews points out that enhancement methods serve as a preprocessing step that boosts object detection performance in degraded underwater conditions. These gains carry over to real-world applications for example, a prototype AUV vision system that can jointly clean and analyze coral reef images in real time. A system like this enables accurate reef survey and biodiversity assessment. By mitigating water-induced blur and color loss, underwater image enhancement has been shown to significantly improve the reliability of vision-based detection, tracking, navigation and 3D mapping. This is a benefit for both industrial robotics ROVs/AUVs and environmental monitoring namely coral and species surveys \cite{jyothimurugan2025efficient}. 

Despite significant progress in UIE, there are many fundamental limitations that remain unresolved. Traditional model-free and physical-model-based methods rely heavily on handcrafted priors and simplified assumptions of underwater light propagation \cite{hummel1975image, ancuti2017color, jin2001contrast, peng2018multi, drews2013transmission, berman2020underwater, peng2017underwater, yuan2020underwater}. While these approaches can improve visibility under specific conditions, they often fail to generalize across diverse underwater environments and are highly sensitive to parameter tuning. Additionally, inaccuracies in estimating scene depth, attenuation coefficients, and background illumination frequently lead to color overcorrection, halo artifacts, and loss of fine details.

Deep learning-based UIE methods \cite{ren2022reinforced, shen2023udaformer, goodfellow2020generative, fu2022uncertainty, li2020underwater, li2019underwater, li2021underwater, liu2019mlfcgan, jahidul2019fast, naik2021shallow, han2023uiegan, yuan2021tebcf, chen2021mffn} address some of these limitations by learning end-to-end mappings from degraded to enhanced images. However, most CNN- and GAN-based approaches are inherently deterministic, producing a single enhancement output for a given input image \cite{shen2023udaformer, jiang2022two}. This formulation neglects the intrinsic ambiguity of underwater image enhancment, where multiple visually plausible solutions may exist due to severe wavelength-dependent attenuation and scattering. As a result, such methods often suffer from over-smoothing, unstable color reproduction, and poor generalization under extreme degradation conditions. GAN-based methods \cite{fabbri2018enhancing}, in particular are prone to training instability and hallucinated textures which compromise structural fidelity and color realism.

Recently, diffusion models have emerged as a promising generative paradigm for low-level vision tasks due to their strong generative priors and stable optimization. However, their application to underwater image enhancement remains limited. Existing diffusion-based UIE methods predominantly rely on small-scale datasets, adopt simplistic conditioning strategies such as direct concatenation, and lack explicit mechanisms to address underwater-specific degradations such as wavelength-dependent color loss and spatially varying haze. Furthermore, most prior diffusion-based approaches employ conventional pixel-wise or perceptual losses that inadequately constrain structural consistency and frequency-domain fidelity, leading to suboptimal restoration of fine textures and global color coherence.

The rest of the study is organized as follows: Section \ref{sec:related_work}
reviews the related work. In Section \ref{sec:methodology}, the proposed AquaDiff
is discussed in detail. In Section \ref{sec:experiments}, extensive experiments are
performed to validate the effectiveness and performance of the
proposed AquaDiff method. Finally, Section \ref{sec:conclusion} consolidates this study.

\section{Related Work}
\label{sec:related_work}

\subsection{Traditional Methods for Underwater Image Enhancement}

Early approaches relied on direct pixel manipulation to improve underwater imagery, using histogram processing \cite{hummel1975image}, white balance \cite{liang2021gudcp}, Retinex formulations, and multi-image fusion. Fu et al. \cite{fu2014retinex} addressed color casts through a correction pipeline and a variational Retinex decomposition that separates reflectance from illumination, yielding more natural tones and brighter outputs. Ancuti et al. \cite{ancuti2012enhancing} a fusion framework combining white balance, histogram equalization, custom weighting, and multiresolution blending to deliver results with stronger global contrast, suppressed noise, and preserved fine structures. Li et al. \cite{gao2021underwater} introduced a contrast enhancement strategy grounded in a natural-scene histogram prior, limiting artifacts while keeping computation modest. Zhang et al. \cite{zhang2024underwater} principal-component fusion to merge contrast-enhanced foregrounds with dehazed backgrounds, improving robustness for downstream underwater perception tasks. Despite these gains, model-free methods often need heavy parameter tuning for different scenes, and their outputs can exhibit over-enhancement or over-exposure artifacts.

Physical-model approaches instead estimate the unknown variables in the underwater formation process using priors and simplifying assumptions. Building on image dehazing advances, early work \cite{ancuti2012enhancing} repurposed the Dark Channel Prior (DCP) \cite{xu2016single} originally devised for atmospheric haze, to enhance underwater visibility yet the markedly different light transport in water often leaves DCP-based outputs visually unsatisfying. Subsequent underwater-specific adaptations \cite{peng2018generalization,lu2015contrast, zhao2015deriving} sought to remedy this. Peng et al. \cite{peng2017underwater} extended DCP by inferring ambient light from depth-dependent color variations and applying adaptive color correction to alleviate color casts. Akkaynak et al. \cite{akkaynak2018revised} revised the widely used Jaffe-McGlamery model \cite{mcglamery1980computer, jaffe2002computer}, showing that direct and backscattered components follow distinct attenuation coefficients governed by factors beyond basic water optics, though the approach relies on specialized instruments to capture precise parameters. Sea-thru \cite{akkaynak2019sea} optimized this revised model, estimating range-dependent attenuation to reduce unknowns and improve reconstruction fidelity. Nonetheless, such model-based techniques hinge on accurate parameter estimation and scene properties; mis-specification can severely degrade results, and the priors themselves are often tailored to specific scenarios, limiting adaptability across varied underwater conditions.

\subsection{Deep Learning-based Methods for Underwater Image Enhancment}

Deep learning has dramatically accelerated underwater image enhancement, with many architectures learning end-to-end mappings from degraded inputs to cleaner outputs. Although such models extract powerful features, their performance often stalls when training data are scarce, limiting robustness in difficult conditions \cite{li2021underwater}. To counter this, recent work injects domain priors into neural networks. Ucolor \cite{li2021underwater} feeds underwater transmission cues into a CNN decoder so the network focuses on severely degraded areas. TAFormer \cite{li2024taformer} leverages transmission maps to supply inductive bias missing from Swin Transformers trained on limited samples. Guo et al \cite{zhang2022underwater} design a color-balance prior that steers hybrid scene-level feature extraction to reduce chromatic distortion; PUGAN \cite{cong2023pugan} a learnable physical model within a GAN, combining physics-based reasoning with data-driven adaptation. Anchored by these priors, the resulting systems generalize more consistently across standard benchmarks.

\subsection{Diffusion Models for Underwater Image Enhancement}

Diffusion models have drawn wide interest for their capacity to synthesize high-quality images by learning rich diffusion priors \cite{ho2020denoising}. They have achieved strong results in low-level vision tasks \cite{nichol2021improved, ho2022cascaded, rombach2022high}, which motivates their application to underwater enhancement \cite{mardani2023variational, ozdenizci2023restoring, kawar2022denoising}. Many diffusion-based UIE frameworks use prior knowledge as conditional guidance.

As diffusion models advance \cite{sohl2015deep}, researchers are applying them to image generation and enhancement. Sohl-Dickstein et al. introduced the diffusion model, a generative approach based on nonequilibrium thermodynamics. It corrupts data by progressively adding noise via a Markov chain, then reconstructs it by progressively removing noise. Despite a solid theoretical foundation, early limitations show unstable training and slow sampling which limits its adoption.

To address these limitations, DDPM \cite{ho2020denoising} presents a streamlined diffusion approach that employs variational inference for modeling and uses reparameterization for sampling. DDPM is a deep generative method that learns the data distribution via forward and reverse diffusion. The forward process incrementally adds Gaussian noise to images, increasing blur and randomness until it approaches an isotropic Gaussian. The reverse process reconstructs the original data from a random Gaussian by incrementally removing noise. This requires a neural network to approximate the conditional probability at each step which predicts the previous state given the current state. DDPM shows stable training and strong image generation results.

Later, Nicholas et al. \cite{nichol2021improved} introduced improved DDPMs (IDDPM), enhancing DDPM to achieve competitive log-likelihoods while preserving high sample quality. Saharia et al. \cite{saharia2022image} two diffusion-based image synthesis methods: Image super-resolution via iterative refinement (SR3) and cascaded diffusion models for high-fidelity image generation (CDM) \cite{ho2022cascaded}. SR3 is a super-resolution diffusion model that generates high-resolution images from low-resolution inputs, starting from pure noise. CDM is a conditional diffusion model that uses category labels as conditions to generate images of the specified category from pure noise.

Diffusion models are now used for image reconstruction tasks \cite{rombach2022high, mardani2023variational}. Kawar et al. \cite{kawar2022denoising} introduced denoising diffusion restoration models (DDRMs) for image recovery.

\begin{figure*}[t]
    \centering
        \includegraphics[width=\textwidth]{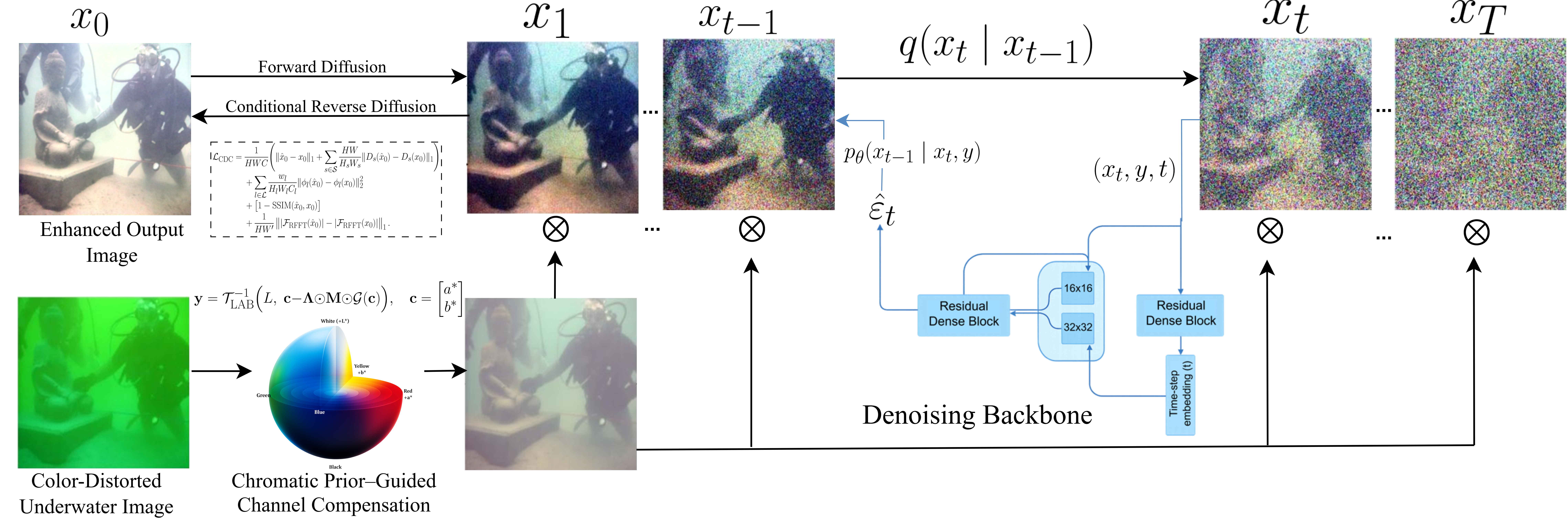}
            \caption{Overview of the proposed AquaDiff framework.The forward diffusion process progressively corrupts the clean reference image $x_0$ into a noisy latent representation $x_t$. During the reverse diffusion process, a denoiser receives the noisy image $x_t$, the chromatic prior--guided conditioning image $y$, and the diffusion timestep $t$, and predicts the noise component $\hat{\varepsilon}_t$.Cross-attention conditioning, denoted as $\otimes\,\mathrm{Cross\mbox{-}Att}(x_t, y)$, is employed to effectively fuse structural information from $x_t$ with chromatic prior guidance from $y$. The predicted noise is then used in the diffusion sampling step to obtain the refined image $x_{t-1}$. This iterative denoising process is repeated across timesteps to progressively recover the enhanced underwater image.
            }
                \label{fig:AquaDiff_architecture}
                \end{figure*}

This variational inference framework uses a pretrained denoising diffusion generative model as a prior for natural images and integrates it with linear measurement methods to approximate a posterior distribution, enabling efficient image generation. DDRMs can be applied to super-resolution, deblurring, restoration, and colorization, producing realistic and diverse results with broad potential.

Lu et al. \cite{lu2023speed} first proposed an UIE method based on DDPM (UW-DDPM), using two U-Net networks for denoising and distribution transformation, improving underwater image quality. However, UW-DDPM was validated on a limited dataset, which limits claims about generalizability. Additionally, Lu et al. \cite{lu2023speed} proposed an accelerated and fused DDPM variant that speeds up inference by modifying the initial sampling distribution and reducing denoising iterations. During diffusion, the degraded and reference images are fused, improving enhanced image quality and avoiding poor quality and color deviation issues that can occur with direct conditional DDPM usage.

In summary, there exists a clear research gap in designing a diffusion-based UIE framework that (i) explicitly incorporates underwater-specific color compensation priors, (ii) employs an effective conditional fusion strategy capable of guiding the denoising process across diffusion timesteps, and (iii) enforces cross-domain consistency to jointly preserve color fidelity, structure, and fine details. Addressing these challenges is essential for achieving robust and perceptually faithful underwater image enhancement in real-world conditions.

To address the aforementioned limitations, we propose AquaDiff, a diffusion-based underwater image enhancement framework specifically designed to correct wavelength-dependent color distortion while preserving structural and perceptual fidelity. The main contributions of this study are summarized as follows:

\begin{itemize}

    \item  We design an enhanced denoising network that combines residual dense blocks, skip connections, and multi-resolution attention modules to capture both global color context and local structural details.

    \item We employ cross-attention mechanisms in AquaDiff to dynamically fuse the noisy intermediate state and the conditioning image at each diffusion step to leverage structural and chromatic cues based on the noise level for improved color correction and detail recovery.

    \item  We propose a novel cross-domain consistency loss that effectively mitigates common diffusion artifacts and ensures visually coherent enhancement results.

    \item We conduct experiments on multiple challenging underwater datasets, including TEST-U90, U45, S16,and C60. Quantitative and qualitative results demonstrate that AquaDiff consistently achieves high results in terms of  color fidelity and competitive overall image quality compared to state-of-the-art traditional, CNN-, GAN-, and diffusion-based methods.
\end{itemize}

\section{The Proposed AquaDiff Model}
\label{sec:methodology}

We propose AquaDiff, a diffusion-based framework specifically designed for underwater image enhancement inspired by \cite{guan2023diffwater}. The diffusion framework employed in AquaDiff operates through a two-stage mechanism: a forward diffusion process that progressively corrupts the clean image with Gaussian noise, and a reverse diffusion process that learns to iteratively denoise and recover the original image as shown in Fig.~\ref{fig:AquaDiff_architecture}. 

In the forward diffusion process, Gaussian noise is progressively introduced to the reference image $x_0$ through a Markov chain, characterized by the transition distribution $q(x_t | x_{t-1})$. This process iteratively corrupts the image over $T$ timesteps, ultimately transforming $x_0$ into a pure noise sample $x_T$ that follows a standard Gaussian distribution.

Conversely, in the reverse diffusion process, the model learns to iteratively denoise and reconstruct the clean image from the noisy state $x_T$. Over $T$ denoising steps, the process progressively refines the image by leveraging the conditional information provided by the color-compensated degraded image $y$. At each timestep $t$, the model utilizes the learned conditional distribution $p_{\theta}(x_{t-1} | x_t, y)$ to predict the previous less-noisy state $x_{t-1}$, ultimately recovering the target distribution $p(x_0 | y)$ through iterative sampling.

To enhance the clarity and quality of underwater imagery, the AquaDiff architecture employs a denoising network based on the SR3 framework ~\cite{saharia2022image}. As illustrated in Fig.~\ref{fig:AquaDiff_architecture}, the denoising model adopts a U-Net backbone architecture incorporating three residual blocks for effective feature extraction and reconstruction.

The network architecture is configured with the following specifications. The initial layer utilizes 64 channels as the base channel width, providing a foundation for feature representation. The network depth is progressively expanded through channel multipliers of $\{1, 2, 4, 8, 16\}$, enabling multi-scale feature extraction at different resolution levels. This hierarchical structure allows the model to capture both fine-grained details and global contextual information essential for underwater image restoration.

To further improve feature representation and extraction capabilities, we incorporate architectural enhancements to the standard U-Net backbone. Specifically, we adopt dense skip connections inspired by U-Net++ architecture, which facilitate better feature fusion between encoder and decoder pathways at multiple resolution levels. Additionally, we replace standard residual blocks with residual dense blocks that incorporate dense connections within each block. This enables richer feature extraction and more expressive representations that can better capture complex degradation patterns in underwater imagery.

To further enhance this capability, we incorporate multi-resolution attention mechanisms at spatial resolutions of $16 \times 16$ and $32 \times 32$ feature map dimensions. This multi-scale attention design enables the model to establish long-range dependencies across the image while preserving local structural details. This is particularly crucial for addressing global color cast issues and local texture preservation challenges inherent in underwater imagery. The attention mechanism operates by computing pairwise relationships between all spatial locations at each resolution level. This allows the model to propagate information across distant regions of the image which further facilitates more consistent color correction throughout the image.

The input preparation process involves two key steps to condition the denoising operation. First, the degraded underwater image undergoes color compensation preprocessing.Subsequently, instead of simple channel concatenation, we employ cross-attention layers to enable more sophisticated fusion between the noisy intermediate state $x_t$ and the color-compensated degraded image $y$. The cross-attention mechanism allows the model to dynamically weight and combine features from the conditioning image based on their relevance to the current denoising step. Specifically, at each timestep $t$, the cross-attention layer computes attention weights that determine how much information from the degraded image $y$ should influence the denoising of the noisy state $x_t$. This adaptive fusion mechanism enables the model to leverage structural information from the degraded image more effectively, particularly in regions where the noise level is high and the original image structure is less discernible. The cross-attention formulation can be expressed as:
\begin{equation}
\text{CrossAtt}(x_t, y)
=
\text{Softmax}\!\left(
\frac{Q(x_t) K(y)^{\top}}{\sqrt{d_k}}
\right)
V(y)
\label{eq:cross_attention}
\end{equation}
where $Q(x_t)$, $K(y)$, and $V(y)$ denote the query, key, and value
projections derived from the noisy latent state $x_t$ and the conditioning image $y$, respectively, and $d_k$ is the dimensionality of the key vectors.This forms a composite input that guides the iterative denoising process. This conditioning mechanism enables the model to leverage both the structural information from the degraded image and the noise schedule information embedded in $x_t$ to progressively recover the clean image.

The following sections provide more details of the AquaDiff framework. We first detail the forward diffusion procedure, which defines the noise corruption process, followed by an explanation of the training methodology for the denoising model $f_{\theta}$ and its application during the inference phase for underwater image enhancement.

\subsection{Chromatic Prior–Guided Channel Compensation}
\label{sec:preprocessing}

Underwater imaging presents significant challenges as a result of the complex interaction of light with aquatic environments. The degradation of underwater images primarily results from three fundamental physical processes, namely absorption, scattering, and reflection of optical signals as they propagate through water. These phenomena collectively manifest as several visual artifacts, including color cast, reduced sharpness, and diminished contrast. These phenomena substantially impede the effectiveness of automated computer vision systems. The severity of these degradations is further influenced by environmental parameters such as water depth, turbidity levels, and temperature variations.

To systematically address these challenges, a physics-based mathematical model characterizes the underwater image formation process. The degradation model for underwater imaging can be expressed as ~\cite{zhang2023metaue}:

\begin{equation}
I_{\lambda}(x) = B_{\lambda}(x)e^{-\eta \cdot d(x)} + J_{\lambda}(x)\left(1 - e^{-\eta \cdot d(x)}\right)
\label{eq:underwater_degradation}
\end{equation}
where $\lambda \in \{R, G, B\}$ denotes the color channel index corresponding to red, green, and blue wavelengths, respectively. In this formulation, $I_{\lambda}(x)$ represents the observed degraded underwater image at pixel location $x$, $B_{\lambda}(x)$ corresponds to the true clean image that would be captured in ideal conditions, $J_{\lambda}(x)$ denotes the background radiance or ambient light component, $\eta$ is the wavelength-dependent attenuation coefficient that quantifies light absorption, and $d(x)$ represents the distance between the camera and the scene point at location $x$.

The observed image $I_{\lambda}(x)$ emerges from the combined effects of water absorption and light scattering mechanisms. The first term $B_{\lambda}(x)e^{-\eta \cdot d(x)}$ captures the exponential attenuation of the direct signal from the scene where the exponential decay factor depends on both the attenuation coefficient and the propagation distance. The second term $J_{\lambda}(x)(1 - e^{-\eta \cdot d(x)})$ accounts for the backscattered ambient light that contributes to the overall image formation. Suspended particles and impurities in water scatter incident light in both forward and backward directions. This leads to reduced visibility and clarity. Furthermore, different wavelengths experience differential attenuation rates with longer wavelengths particularly red being absorbed more rapidly than shorter wavelengths such as blue. This results in the characteristic color distortion and low contrast observed in underwater imagery. These degradation effects are quantitatively described by the background brightness $J_{\lambda}(x)$, the attenuation coefficient $\eta$, and the distance function $d(x)$.

Motivated by this physical model of underwater image degradation, a conventional preprocessing technique which is the 3-channel compensation (3C) \cite{ancuti2019color} was used to mitigate color distortion and facilitate the recovery of true color information and fine details in underwater images. This method operates on the principle of reconstructing attenuated color channels by making use of complementary chromatic information from the Lab color space.

The compensation process is mathematically formulated through the following equations:

\begin{equation}
I_a^c(x) = I_a(x) - \kappa \cdot M(x) \cdot G[I_a(x)]
\label{eq:channel_a_compensation}
\end{equation}

\begin{equation}
I_b^c(x) = I_b(x) - \lambda \cdot M(x) \cdot G[I_b(x)]
\label{eq:channel_b_compensation}
\end{equation}
where $I_a^c(x)$ and $I_b^c(x)$ represent the compensated chromatic channels in the Lab color space, $I_a(x)$ and $I_b(x)$ denote the original chromatic channels of the degraded underwater image, $\kappa$ and $\lambda$ are compensation parameters that control the degree of correction for the respective channels, $G[\cdot]$ denotes the Gaussian blur operator, and $M(x)$ is a spatially-varying mask that modulates the compensation intensity.

The channel-wise compensation defined in Eqs. (2)–(3) can be expressed compactly in vector form as follows:

\begin{equation}
\mathbf{y} =
\mathcal{T}_{\mathrm{LAB}}^{-1}
\Big(
L,\;
\mathbf{c}
-
\mathbf{\Lambda}
\odot
\mathbf{M}
\odot
\mathcal{G}(\mathbf{c})
\Big),
\quad
\mathbf{c} =
\begin{bmatrix}
a^* \\
b^*
\end{bmatrix}
\label{eq:unified_chromatic_compensation}
\end{equation}
where the compensation parameters $\kappa$ and $\lambda$ determine the strength of the correction applied to each chromatic channel. On the basis of empirical analysis from the 3C method, optimal performance is usually achieved when both parameters are set to approximately 0.7.

The mask $M(x)$ serves a critical role in preventing over-compensation in regions of high brightness, particularly near light sources. The mask is constructed by thresholding the grayscale representation of the input image where pixels with average brightness exceeding 0.85 are assigned a value of zero and all other pixels are set to one. To ensure smooth transitions and avoid artifacts at mask boundaries, the binary mask is subsequently smoothed using a Gaussian filter.

The complete 3-channel compensation pipeline proceeds through three sequential steps. The first step is color space transformation. The degraded underwater RGB image is first converted to the Lab color space, which separates luminance information that is L channel from chromatic information which are the a and b channels. The chromatic channels $I_a(x)$ and $I_b(x)$ are extracted for subsequent processing. The second step is mask generation. A grayscale representation of the input RGB image is generated, from which the compensation mask $M(x)$ is constructed through thresholding and Gaussian smoothing operations as described above. The third step is channel compensation. For each chromatic channel, the compensation is performed by subtracting the element-wise product of the Gaussian-blurred channel and the mask $M(x)$, scaled by the respective compensation parameter. This means that each chromatic channel is compensated by subtracting a masked, Gaussian-smoothed version of itself, which suppresses dominant color casts while preserving local structure. Specifically, $I_a(x)$ is compensated using parameter $\kappa$ according to Equation~\eqref{eq:channel_a_compensation}, while $I_b(x)$ is compensated using parameter $\lambda$ according to Equation~\eqref{eq:channel_b_compensation}. The compensated channels $I_a^c(x)$ and $I_b^c(x)$ are then combined with the original L channel and converted back to RGB color space to produce the color-corrected image.

The UIE task fundamentally differs from the standard reverse diffusion process employed in Denoising Diffusion Probabilistic Models (DDPM). Unlike DDPM, which generates images from pure noise, the primary objective of UIE is to learn an iterative mapping that transforms a degraded underwater image into a high-quality reference image. This process aims to approximate the conditional probability distribution $p(x_0 | y)$, where $y$ represents the color-compensated underwater degraded image which was obtained through the preprocessing step described in Section~\ref{sec:preprocessing} and $x_0$ denotes the target underwater reference image.

\subsection{Forward Diffusion Process with Progressive Noise Injection}
\label{sec:Forward Diffusion Process with Progressive Noise Injection}

The forward diffusion process is formulated as a Markov chain $q$~\cite{ho2020denoising} that gradually introduces Gaussian noise into the underwater reference image $x_0$ over $T$ discrete timesteps. It ultimately transforms it into a pure noise sample $x_T$ that follows a standard Gaussian distribution. This process is governed by a predefined variance schedule $\{\beta_1, \beta_2, \ldots, \beta_T\}$, where each $\beta_t$ controls the noise intensity at timestep $t$.

The joint probability distribution of the entire forward diffusion trajectory, conditioned on the initial clean image $x_0$, is expressed as:

\begin{equation}
q(x_{1:T} | x_0) = \prod_{t=1}^{T} q(x_t | x_{t-1})
\label{eq:forward_joint}
\end{equation}
where each transition in the Markov chain follows a Gaussian distribution:

\begin{equation}
q(x_t | x_{t-1}) = \mathcal{N}(x_t; \sqrt{1 - \beta_t} x_{t-1}, \beta_t I)
\label{eq:forward_transition}
\end{equation}
where $t \in \{1, 2, \ldots, T\}$ denotes the diffusion timestep, and $\beta_t \in (0, 1)$ is a scalar hyperparameter that determines the variance of the Gaussian noise introduced at each iteration. The parameter $\beta_t$ quantifies the noise intensity at timestep $t$, with the noise level gradually increasing as the diffusion process progresses from $t=1$ to $t=T$.

Conversely, the reverse diffusion process operates in the opposite direction, starting from fully Gaussian noise $x_T$ and functioning as a learned denoising mechanism that progressively recovers the clean image. The noise schedule parameter $\beta_t$ serves as a crucial hyperparameter that determines the strength of noise at each step. This enables controlled denoising during the reverse process. To facilitate parameter sharing across different timesteps, the model incorporates time step embeddings that encode the current diffusion step $t$, allowing the denoising network to adapt its behavior based on the noise level. The noise intensity $\beta_t$ is typically configured to increase linearly with the timestep $t$, ensuring a smooth and gradual corruption process.

To enable efficient sampling at arbitrary timesteps without requiring iterative computation through the entire Markov chain, we derive a closed-form expression for $q(x_t | x_0)$. Letting $\alpha_t = 1 - \beta_t$ and defining the cumulative product $\bar{\alpha}_t := \prod_{s=1}^{t} \alpha_s$, the distribution of $x_t$ conditioned on $x_0$ can be expressed as:

\begin{equation}
q(x_t | x_0) = \mathcal{N}(x_t; \sqrt{\bar{\alpha}_t} x_0, (1 - \bar{\alpha}_t) I).
\label{eq:closed_form}
\end{equation}

This closed-form formulation allows direct sampling of $x_t$ from $x_0$ at any timestep $t$ without sequentially applying the forward diffusion steps, significantly improving computational efficiency during training and inference.

For notational simplicity, we introduce $\gamma_t = \bar{\alpha}_t$, allowing Equation~\eqref{eq:closed_form} to be expressed more compactly as:

\begin{equation}
q(x_t | x_0) = \mathcal{N}(x_t; \sqrt{\gamma_t} x_0, (1 - \gamma_t)I)
\label{eq:simplified_closed_form}
\end{equation}
where $\gamma_t$ represents the cumulative product of the noise schedule parameters up to timestep $t$, encapsulating the total amount of noise that has been introduced to the original image.

To ensure numerical stability and bounded variance as the diffusion process progresses, the noise schedule parameter $\beta_t$ is configured to decrease with increasing $t$, maintaining the variance of the random variable within acceptable bounds as $t \to \infty$. Through reparameterization of the closed-form distribution in Equation~\eqref{eq:simplified_closed_form}, we can directly sample the noisy image $x_t$ at any timestep $t$ using the following formulation:

\begin{equation}
x_t = \sqrt{\gamma_t} x_0 + \sqrt{1 - \gamma_t} \varepsilon
\label{eq:reparameterization}
\end{equation}
where $\varepsilon \sim \mathcal{N}({0}, {I})$ follows a standard multivariate Gaussian distribution with zero mean and identity covariance matrix. This reparameterization enables efficient sampling of intermediate noisy states during the forward diffusion process, as $x_t$ can be directly computed from the original image $x_0$ and a randomly sampled noise vector $\varepsilon$, without requiring sequential application of the Markov chain transitions.

The reparameterization formulation in Equation~\eqref{eq:reparameterization} provides a computationally efficient mechanism for generating training samples at arbitrary noise levels, which is essential for training the denoising network. In the subsequent section, we detail the methodology for learning a neural network $f_{\theta}$ that learns to reverse this forward diffusion process, enabling the iterative recovery of clean images from noisy states.

\subsection{Inverse Diffusion by Adding Iterative Refinement of Conditions}

The inference process in AquaDiff is formulated as a reverse Markov process that operates in the opposite direction of the forward diffusion. This process iteratively recovers the clean image from a pure noise state. Unlike the forward process that corrupts the image, the reverse process learns to progressively denoise and restore the target image $x_0$ through $T$ iterative refinement steps, conditioned on the color-compensated degraded image $y$.

The joint probability distribution of the reverse diffusion trajectory, conditioned on the input $y$, is expressed as:

\begin{equation}
p_{\theta}(x_{0:T} | y) = p(x_T) \prod_{t=1}^{T} p_{\theta}(x_{t-1} | x_t, y)
\label{eq:reverse_joint}
\end{equation}
where $p(x_T)$ represents the prior distribution at the final timestep, and $p_{\theta}(x_{t-1} | x_t, y)$ denotes the learned conditional distribution that governs the transition from the noisy state $x_t$ to the less noisy state $x_{t-1}$, given the conditioning information $y$.

The prior distribution $p(x_T)$ is defined as a standard Gaussian:

\begin{equation}
p(x_T) = \mathcal{N}(x_T | 0, I).
\label{eq:prior}
\end{equation}

This formulation allows the reverse process to commence from pure Gaussian noise. This ensures that the initial state $x_T$ is independent of the target image and can be sampled randomly.

The conditional distribution $p_{\theta}(x_{t-1} | x_t, y)$ is parameterized as a Gaussian distribution:

\begin{equation}
p_{\theta}(x_{t-1} | x_t, y) = \mathcal{N}(x_{t-1} | \mu_{\theta}(y, x_t, t), \sigma_t^2 I)
\label{eq:reverse_conditional}
\end{equation}
where $y$ represents the color-compensated degraded underwater image, $\theta$ denotes the learnable parameters of the denoising network $f_{\theta}$, $\mu_{\theta}(y, x_t, t)$ is the predicted mean of the distribution and $\sigma_t^2$ is the variance parameter at timestep $t$.

The inference process is based on the isotropic Gaussian conditional formulation~\cite{saharia2022image} which provides a principled framework for the reverse diffusion. The process initiates from a randomly sampled Gaussian noise $x_T \sim \mathcal{N}(0, I)$ and progressively reverses the forward diffusion through $T$ iterative refinement steps. At each timestep $t$, the model utilizes the learned conditional distribution $p_{\theta}(x_{t-1} | x_t, y)$ to predict the previous less-noisy state, ultimately recovering the target image $x_0$.

Under certain regularity conditions, the optimal reverse process can be well-approximated by a Gaussian distribution~\cite{saharia2022image}. This makes the Gaussian parameterization in Equation~\eqref{eq:reverse_conditional} a natural and effective choice. To ensure that the initial state $x_T$ conforms to the prior distribution $p(x_T) = \mathcal{N}(x_T | 0, I)$, the noise schedule is configured such that $1 - \gamma_T$ is sufficiently large, allowing the sampling process to commence from random Gaussian noise.

Following the parameterization approach of Ho et al.~\cite{ho2020denoising}, we define the posterior mean and variance for the reverse process. The mean parameter $\mu_{\theta}(y, x_t, t)$ is expressed as:

\begin{equation}
\mu_{\theta}(y, x_t, t) = \frac{\sqrt{\gamma_{t-1}(1 - \alpha_t)}}{1 - \gamma_t} x_0 + \frac{\sqrt{\alpha_t}(1 - \gamma_{t-1})}{1 - \gamma_t} x_t
\label{eq:mean_parameterization}
\end{equation}
where $\alpha_t = 1 - \beta_t$ and $\gamma_t = \prod_{s=1}^{t} \alpha_s$ as defined in the forward process. The variance parameter is given by:

\begin{equation}
\sigma_t^2 = \frac{(1 - \gamma_{t-1})(1 - \alpha_t)}{1 - \gamma_t}.
\label{eq:variance_parameterization}
\end{equation}

The core learning objective is to train a denoising score model $f_{\theta}$ that estimates the noise component $\varepsilon$ given a noisy image $x_t$, the conditioning image $y$, and the diffusion timestep $t$. During training, the network $f_{\theta}(y, x_t, t)$ learns to predict the noise $\varepsilon$ that was added to the clean image $x_0$ to produce $x_t$.

By substituting the predicted noise $f_{\theta}(y, x_t, t)$ for $\varepsilon$ in the reparameterization equation (Equation~\eqref{eq:reparameterization}) and rearranging terms, we obtain an approximation of the target clean image:

\begin{equation}
\hat{x}_0 = \frac{1}{\sqrt{\gamma_t}} \left( x_t - \sqrt{1 - \gamma_t} f_{\theta}(y, x_t, t) \right).
\label{eq:target_approximation}
\end{equation}

This formulation enables the model to iteratively refine the image estimate by predicting and removing the noise component at each timestep, progressively recovering the clean image $x_0$ from the noisy state $x_t$ through the learned denoising function $f_{\theta}$.

By solving Equations~\eqref{eq:mean_parameterization} and~\eqref{eq:target_approximation}, the mean parameter $\mu_{\theta}(y, x_t, t)$ of the conditional distribution $p_{\theta}(x_{t-1} | x_t, y)$ in Equation~\eqref{eq:reverse_conditional} can be expressed in a simplified form:

\begin{equation}
\mu_{\theta}(y, x_t, t) = \frac{1}{\sqrt{\alpha_t}} \left( x_t - \frac{1 - \alpha_t}{\sqrt{1 - \gamma_t}} f_{\theta}(y, x_t, t) \right)
\label{eq:simplified_mean}
\end{equation}
where $\alpha_t = 1 - \beta_t$ and $\gamma_t = \prod_{s=1}^{t} \alpha_s$ denote the variance parameters that govern noise injection in the forward diffusion process, as defined in Section~\ref{sec:Forward Diffusion Process with Progressive Noise Injection}.

Following this parameterization, each refinement step in the reverse diffusion process samples the next state $x_{t-1}$ according to:

\begin{equation}
\begin{aligned}
x_{t-1} \sim\;& \frac{1}{\sqrt{\alpha_t}}
\left(
x_t - \frac{1 - \alpha_t}{\sqrt{1 - \gamma_t}} 
f_{\theta}(y, x_t, t)
\right) \\
&+ \sqrt{\frac{(1 - \gamma_{t-1})(1 - \alpha_t)}{1 - \gamma_t}} \,
\varepsilon_t
\end{aligned}
\label{eq:sampling_step}
\end{equation}
where $\varepsilon_t \sim \mathcal{N}(0, I)$ is a standard Gaussian noise vector sampled independently at each timestep. This sampling formulation is analogous to a Langevin dynamics step, where the denoising network $f_{\theta}$ provides an estimate of the data log-density gradient, guiding the iterative refinement toward the target distribution.

Throughout the $T$-step inverse diffusion process, the color channel-compensated image $y$ obtained through the 3-channel compensation operation described in Section~\ref{sec:preprocessing} is consistently applied as a conditioning signal at each refinement step. When the degraded image $\tilde{y}$ is used directly as the input condition, it provides rich structural information to the neural network, enabling more effective denoising.

The contour features and structural patterns of the original clean image $x_0$ serve as valuable priors that assist the neural network in accurately predicting the noise component $\varepsilon$ and recovering the original image $x_0$. However, due to the severe degradation characteristics inherent in underwater imagery, directly using the raw degraded image $\tilde{y}$ may not provide sufficient information for effective restoration.

To address this limitation, we employ a preliminarily optimized underwater degraded image $y$, obtained through color compensation preprocessing. This color-compensated image $y$ serves as an enhanced input condition that provides prior knowledge and structural guidance to the neural network, enabling more effective utilization of the noise variance schedule $\beta_t$. By incorporating this conditioning information, the model can more accurately predict noise at each timestep and reduce color distortion, thereby enhancing both the effectiveness and quality of the image restoration process.

In our model AquaDiff, the inverse diffusion process is carried by a neural network that is employed to predict the noise component at each refinement step. This enables the iterative recovery of the clean image from the noisy state. To manage computational complexity while maintaining denoising effectiveness, we adopt a U-Net architecture which has demonstrated strong performance in diffusion-based image enhancement tasks.

The denoising network $f_{\theta}$ takes as input the current noisy image state $x_t$ and the diffusion timestep $t$ and produces as output an estimate of the noise component $\varepsilon$ that was added to the clean image. Formally, the network mapping can be expressed as:

\begin{equation}
f_{\theta}: (x_t, y, t) \mapsto \hat{\varepsilon}
\label{eq:network_mapping}
\end{equation}
where $y$ represents the color-compensated degraded image used for conditioning, and $\hat{\varepsilon}$ denotes the predicted noise estimate.

During the training phase, the neural network is utilized in a single forward pass to predict the noise component $\varepsilon$ from a randomly sampled noisy state $x_t$ and the corresponding timestep $t$. The network parameters $\theta$ are optimized to minimize the discrepancy between the predicted noise $\hat{\varepsilon}$ and the actual noise $\varepsilon$ that was introduced during the forward diffusion process.

In contrast, during the reverse diffusion process for inference, the network must be evaluated $T$ times, where $T$ is typically set to approximately 1000 iterations. In each iteration, the neural network predicts the noise component, and a deterministic update formula (as specified in Equation~\eqref{eq:sampling_step}) is employed to refine the image estimate $x_t$ toward the clean image $x_0$.

While the computational load during training is relatively manageable, as it requires only a single network evaluation per training sample, the inference process becomes computationally intensive and time-consuming. This computational burden arises from the necessity of performing $T$ sequential network evaluations, each requiring a full forward pass through the U-Net architecture. This challenge is not unique to the proposed AquaDiff method but represents a fundamental characteristic shared by all diffusion-based generative models which trade off inference efficiency for high-quality image generation capabilities.

\subsection{Cross-Domain Consistency Loss}

Training diffusion models for underwater image enhancement requires a carefully designed loss function capable of simultaneously addressing multiple degradation factors, including color distortion, loss of fine details, and structural degradation caused by wavelength-dependent light absorption and scattering in underwater environments. To this end, we propose cross-domain consistency loss (CDC) that jointly enforces consistency across pixel, multi-scale, perceptual, structural, and frequency domains for underwater image enhancement.

The cross-domain consistency loss is defined as:
\begin{equation}
\begin{split}
\mathcal{L}_{\mathrm{CDC}}
=\;&
\frac{1}{HWC}
\Bigg(
\|\hat{x}_0 - x_0\|_1
+
\sum_{s \in \mathcal{S}}
\frac{HW}{H_s W_s}
\|D_s(\hat{x}_0) - D_s(x_0)\|_1
\Bigg)
\\
&+
\sum_{l \in \mathcal{L}}
\frac{w_l}{H_l W_l C_l}
\|\phi_l(\hat{x}_0) - \phi_l(x_0)\|_2^2
\\
&+
\bigl[1-\mathrm{SSIM}(\hat{x}_0,x_0)\bigr]
\\
&+
\frac{1}{HW'}
\left\|
\begin{aligned}
|\mathcal{F}_{\mathrm{RFFT}}(\hat{x}_0)|
-
|\mathcal{F}_{\mathrm{RFFT}}(x_0)|
\end{aligned}
\right\|_1 .
\end{split}
\end{equation}
where $H$, $W$, and $C$ denote the height, width, and number of channels of the input image, respectively. The operator $D_s(\cdot)$ represents spatial downsampling at scale $s \in \mathcal{S} = \{0.5, 0.25\}$, while $H_s$ and $W_s$ denote the corresponding spatial dimensions at scale $s$. The symbol $\phi_l(\cdot)$ denotes feature activations extracted from the $l$-th layer of a pre-trained VGG-19 network, with $\mathcal{L} = \{2, 7, 16\}$ indicating the selected feature layers and $w_l$ representing layer-specific weighting coefficients.

The first term corresponds to a pixel-wise $\ell_1$ reconstruction loss computed at the original image resolution, which enforces fundamental fidelity between the predicted enhanced image $\hat{x}_0$ and the ground-truth target image $x_0$. To further promote structural consistency across spatial resolutions, the second term introduces a multi-scale $\ell_1$ loss that measures reconstruction errors at progressively coarser scales. This multi-scale formulation encourages the preservation of both local details and global image structures, which is particularly important in underwater scenarios where degradations are spatially varying.

The perceptual loss term operates in feature space rather than pixel space, leveraging deep semantic representations from the VGG-19 network to capture high-level perceptual similarities between images. By minimizing feature discrepancies across multiple layers, this component helps restore textures, contrast, and semantic consistency that are not adequately captured by pixel-wise losses alone.

Structural similarity is enforced using the SSIM-based loss term, which evaluates the similarity between $\hat{x}_0$ and $x_0$ in terms of luminance, contrast, and structural information. This term effectively reduces structural distortions and mitigates artifacts that commonly arise during image enhancement. Finally, the frequency-domain loss compares the magnitude spectra of the predicted and target images using the real-valued fast Fourier transform. By constraining differences in the frequency domain, this component encourages the recovery of high-frequency details such as edges and fine textures that are often attenuated in underwater imagery due to scattering and turbidity.

Together, these components form a unified hybrid loss function that jointly optimizes pixel-level accuracy, perceptual quality, structural consistency, and frequency-domain fidelity. This enables robust and visually coherent underwater image enhancement within the diffusion modeling framework.

\section{Experiments}
\label{sec:experiments}

\subsection{Datasets}

In this subsection, to ensure evaluation of the proposed AquaDiff method, we use multiple publicly available underwater image datasets that encompass diverse degradation scenarios and imaging conditions. The training phase utilizes two primary datasets: the LSUI dataset~\cite{peng2023u} and the UIEB dataset~\cite{li2019underwater}, which together provide a substantial collection of paired underwater images for learning effective enhancement mappings. The LSUI dataset comprises 5,004 paired underwater images. This dataset offers extensive coverage of various underwater environments, degradation types, and scene compositions. The UIEB dataset~\cite{li2019underwater} contains 890 paired underwater images, with 800 image pairs designated for training purposes, providing additional diversity in underwater imaging conditions and enhancement challenges. These training datasets collectively enable the model to learn robust enhancement strategies across a wide spectrum of underwater degradation patterns, including color cast variations, haze effects, contrast reduction, and detail loss.

\begin{table*}[t!]
\centering
\small
\caption{Quantitative comparison of state-of-the-art UIE methods with AquaDiff (ours) on U45, S16, C60, and TEST-U90 datasets. The best results are highlighted in bold.}
\label{tab:quantitative_results}
\resizebox{\textwidth}{!}{
\begin{tabular}{lcccccc|cc}
\toprule
\multirow{2}{*}{Methods} 
& \multicolumn{2}{c}{U45} 
& \multicolumn{2}{c}{S16} 
& \multicolumn{2}{c}{C60}
& \multicolumn{2}{c}{TEST-U90} \\
\cmidrule(lr){2-3} 
\cmidrule(lr){4-5} 
\cmidrule(lr){6-7}
\cmidrule(lr){8-9}
 & UIQM & UCIQE & UIQM & UCIQE & UIQM & UCIQE & PSNR & SSIM \\
\midrule

UDCP~\cite{drews2013transmission} 
& 3.3019 & 0.4546 
& 1.4920 & 0.4431 
& 2.7336 & 0.3926
& 11.3752 & 0.5156 \\

UIBLA~\cite{peng2017underwater} 
& 2.6307 & 0.4557
& 1.4269 & 0.4042
& 3.0864 & 0.4081
& 15.1351 & 0.6454 \\

UWCNN-typeII~\cite{li2020underwater}
& 3.8481 & 0.3736
& 3.2716 & 0.3698
& 3.3557 & 0.3348
& 13.8223 & 0.6385 \\

Shallow-UWNet~\cite{naik2021shallow}
& 4.0523 & 0.3703
& 3.1680 & 0.3861
& 4.2963 & 0.3348
& 15.9740 & 0.7099 \\

Ucolor~\cite{li2021underwater}
& \best{4.9504} & 0.4461
& 3.5754 & 0.4185
& 4.3337 & 0.3846
& \best{21.0005} & 0.8693 \\

FUnIEGAN~\cite{islam2020fast}
& 4.3502 & 0.3997
& 2.8500 & 0.3955
& 4.2946 & 0.3592
& 16.9189 & 0.7338 \\

MLFcGAN~\cite{liu2019mlfcgan}
& 4.0229 & 0.3908
& 3.3020 & 0.4172
& 4.2923 & 0.3681
& 15.1004 & 0.6573 \\

Water-Net~\cite{li2019underwater}
& 4.8603 & 0.4502
& 3.4989 & 0.4307
& 4.4453 & 0.4416
& 19.9223 & 0.8326 \\

DiffWater~\cite{guan2023diffwater}
& 4.7306 & 0.4624
& \best{4.5159} & 0.4503
& \best{4.6559} & 0.4339
& 20.9721 & \best{0.8951} \\

\midrule
\textbf{AquaDiff (Ours)}
& 4.6097 & \best{0.5390}
& 4.4385 & \best{0.5243}
& 4.3243 & \best{0.5176}
& 20.2494 & 0.8832 \\   
\bottomrule
\end{tabular}
}
\end{table*}

For evaluation purposes, we utilize several independent test datasets that were not included in the training phase. This ensures unbiased assessment of generalization capabilities. The UIEB test set, referred to as TEST-U90, contains 90 paired underwater images that represent diverse underwater scenarios and degradation characteristics. Additionally, we evaluate on the U45 dataset~\cite{li2019fusion}. This dataset is a public underwater image test dataset containing 45 underwater images captured across different scenes. The U45 dataset encompasses images exhibiting various underwater degradation effects, including color cast, low contrast, and haze effects. These conditions make it an ideal benchmark for assessing method robustness. The entire 45 images in the U45 dataset are used as the test set, referred to as U45. Furthermore, we evaluate on the C60 dataset, which consists of 60 challenging underwater images from the UIEB dataset that cannot obtain satisfactory reference images. These are referred to as C60. These challenging cases test the method's capability to handle severely degraded scenarios where ground truth references are unavailable or unreliable. Additionally, we employ the S16 dataset for evaluation, which provides 16 test images with specific degradation characteristics. The SQUID dataset~\cite{berman2021squid} is also utilized for evaluation. This dataset consists of real underwater images captured in natural marine locations, primarily in coastal and near-shore waters, including reefs, seabeds, and harbor-like environments. The images were acquired using optical underwater cameras operated by divers and underwater platforms, rather than controlled laboratory setups. As a result, the dataset reflects authentic in-situ conditions. It includes images affected by depth variation, natural sunlight attenuation, water turbidity, and scattering commonly encountered in real ocean scenes.

\subsection{Implementation Details}

AquaDiff is implemented using PyTorch 1.12.1 and trained on NVIDIA GeForce RTX 3090 GPU. The training process operates on image patches cropped to $256 \times 256$ pixels. This step provides a balance between computational efficiency and sufficient spatial context for learning degradation patterns. The diffusion process employs $T = 2000$ timesteps with the variance schedule of the forward diffusion process configured to increase linearly from $\alpha_1$ to $\alpha_T$ over the range $[10^{-6}, 10^{-2}]$. This follows the schedule proposed in \cite{saharia2022image}.

The model is optimized using the Adam optimizer with hyperparameters $\beta_1 = 0.9$ and $\beta_2 = 0.999$. Training commences with an initial learning rate of $3 \times 10^{-6}$ and proceeds for one million iterations. The batch size is set to 1 which accommodates the memory constraints of the diffusion model architecture while enabling stable gradient updates. This training configuration ensures sufficient exposure to diverse underwater degradation patterns while maintaining computational feasibility.

The training dataset is constructed from two publicly available underwater image enhancement benchmarks. We utilize 5,004 paired underwater images from the LSUI dataset~\cite{peng2023u} and 800 paired images from the UIEB dataset~\cite{li2019underwater}. From the UIEB dataset, 800 image pairs are excluded from training, leaving 90 pairs for validation and testing purposes. From these remaining 90 pairs, three pairs are selected to form the validation set, which is used to monitor training progress and select the optimal model checkpoint. The pretrained model demonstrating the best performance on this validation set is subsequently employed for evaluation on the remaining test sets.

\subsection{Comparison with the State-of-the Art Methods}

To comprehensively evaluate the effectiveness of the proposed AquaDiff method, we conduct extensive comparisons against a diverse set of state-of-the-art underwater image enhancement approaches. The comparative methods encompass both traditional model-free techniques and modern deep learning-based approaches, providing a thorough assessment across different methodological paradigms.

\begin{figure*}[t]
    \centering
    \includegraphics[width=\textwidth]{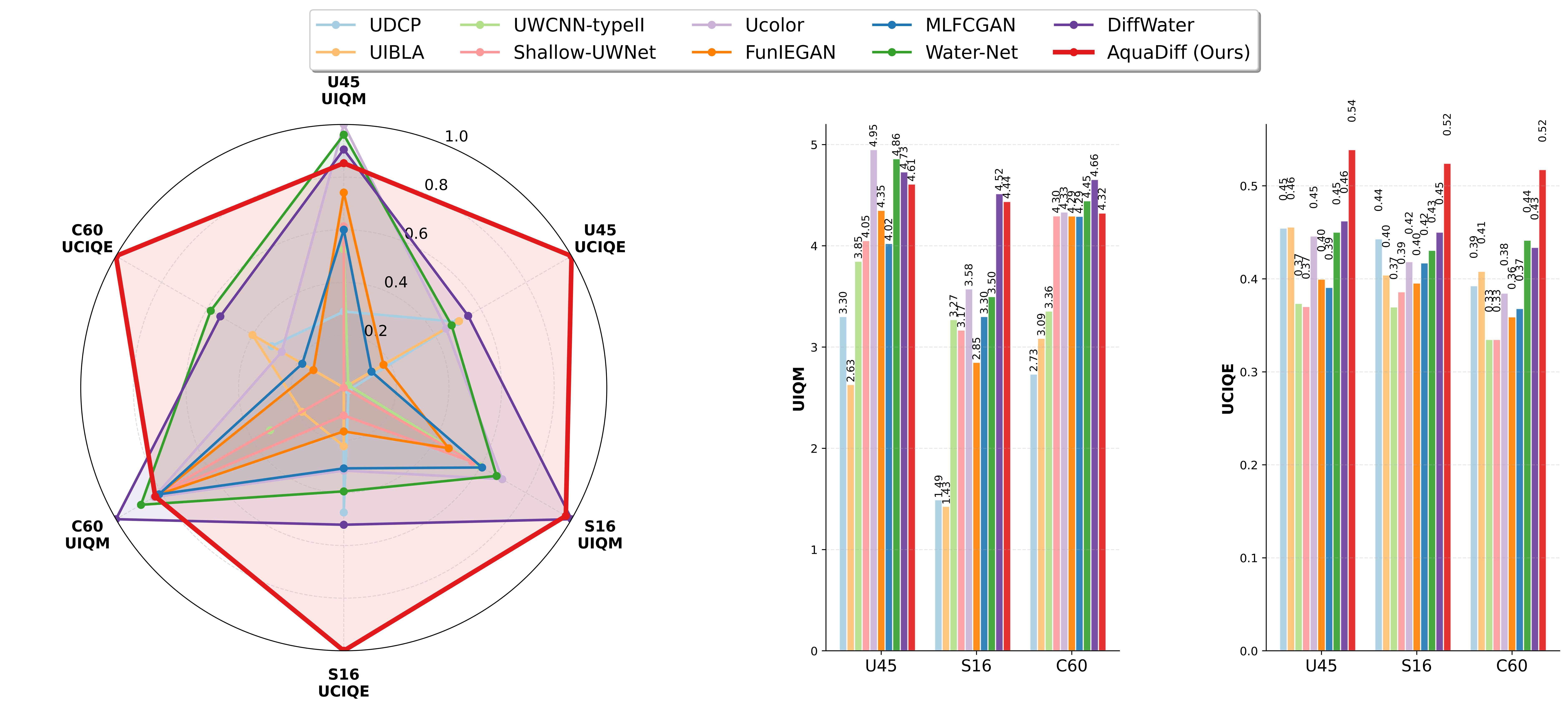}
    \caption{Quantitative comparison of underwater image enhancement methods across the U45, S16, and C60 datasets. The radar plot illustrates performance trends, while the bar charts present absolute UIQM and UCIQE scores. AquaDiff consistently achieves high values, highlighting its capability in restoring color fidelity, contrast, and overall visual quality.}
    \label{fig:AquaDiff_UIQM_UCIQE_radar_bars}
\end{figure*}

Specifically, we compare AquaDiff against two conventional model-free methods: UDCP~\cite{drews2013transmission} and UIBLA~\cite{peng2017underwater}, which serve as baseline references for traditional underwater image enhancement techniques. Additionally, we evaluate against six deep learning-based methods that represent the current state-of-the-art in the field: UWCNN~\cite{li2020underwater}, Water-Net~\cite{li2019underwater}, Ucolor~\cite{li2021underwater}, MLFcGAN~\cite{liu2019mlfcgan}, FUnIEGAN~\cite{islam2020fast}, and Shallow-uwnet~\cite{naik2021shallow}. Among these deep learning approaches, UWCNN, Shallow-uwnet, and Ucolor are CNN-based architectures that leverage convolutional neural networks for feature extraction and image restoration, while MLFcGAN and FUnIEGAN represent GAN-based generative approaches that employ adversarial training paradigms for underwater image enhancement.

To ensure a fair and rigorous evaluation, both the proposed AquaDiff method and all comparative methods are trained on the same training datasets, specifically the LSUI~\cite{peng2023u} and UIEB~\cite{li2019underwater} datasets following identical data preprocessing protocols. Subsequently, all methods are evaluated on independent test datasets that were not included in the training phase, thereby ensuring that the performance comparisons reflect genuine generalization capabilities rather than dataset-specific overfitting. This experimental protocol guarantees that the comparative analysis provides meaningful insights into the relative strengths and limitations of each approach across varying data distributions and underwater imaging conditions.

The experimental comparisons presented in the following sections serve to validate the effectiveness and robustness of the proposed AquaDiff method. Through both quantitative metrics and qualitative visual assessments, we demonstrate that AquaDiff achieves superior performance in addressing key challenges of underwater image enhancement including color distortion correction, detail preservation, and overall visual quality improvement, while maintaining computational efficiency suitable for practical applications.

\subsection{Quantitative Evaluations}

To evaluate the performance of different underwater image enhancement methods, we employ both full-reference and no-reference quality assessment metrics. Full-reference metrics require ground truth reference images for comparison and provide objective measures of enhancement accuracy. Specifically, we utilize Peak Signal-to-Noise Ratio (PSNR)~\cite{a65} and Structural Similarity Index (SSIM)~\cite{a66} as full-reference metrics. PSNR quantifies the pixel-wise difference between enhanced and reference images, providing a measure of reconstruction fidelity, while SSIM evaluates structural similarity by considering luminance, contrast, and structure, offering a more perceptually relevant assessment of image quality.

For scenarios where reference images are unavailable or unreliable, we employ no-reference metrics that assess image quality based solely on the enhanced image characteristics. Specifically, we utilize Underwater Color Image Quality Evaluation (UCIQE)~\cite{a66} and Underwater Image Quality Measure (UIQM)~\cite{a66} as no-reference metrics. UCIQE evaluates underwater image quality by considering colorfulness, contrast, and saturation. 

\begin{figure*}[t]
    \centering
    \includegraphics[width=\textwidth]{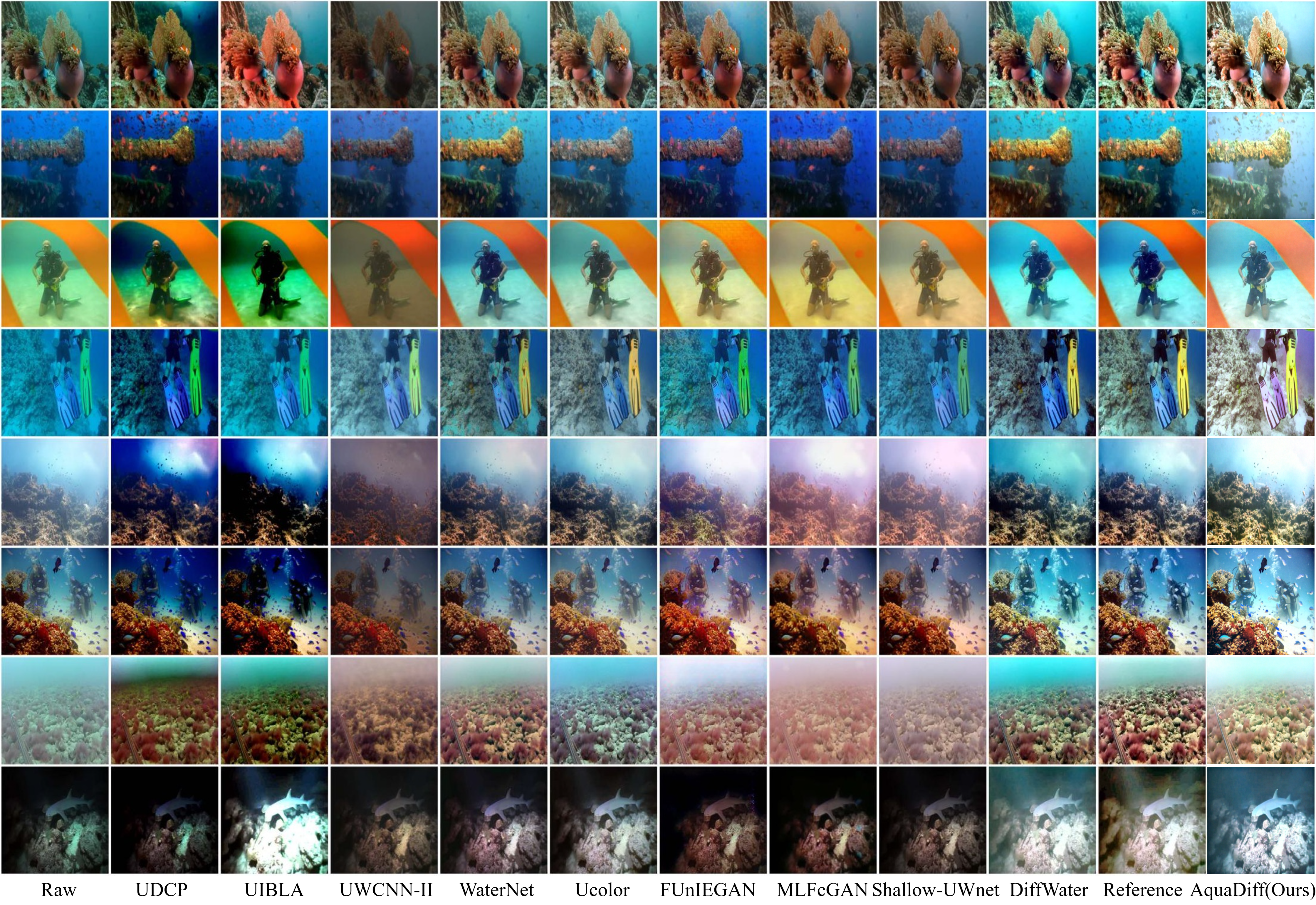}
    \caption{Qualitative comparison of our proposed method AquaDiff with traditional methods namely UDCP~\cite{drews2013transmission}, UIBLA~\cite{peng2017underwater} and deep learning-based methods namely UWCNN~\cite{li2020underwater}, Water-Net~\cite{li2019underwater}, Ucolor~\cite{li2021underwater}, MLFcGAN~\cite{liu2019mlfcgan}, FUnIEGAN~\cite{islam2020fast}, Shallow-uwnet~\cite{naik2021shallow}, DiffWater~\cite{guan2023diffwater} for underwater image enhancment on the U-90 dataset.}
    \label{fig:U90}
\end{figure*}

UIQM incorporates multiple underwater-specific quality factors, including colorfulness, sharpness, and contrast. These metrics collectively provide an evaluation of both quantitative accuracy and perceptual quality that enables a thorough assessment of method performance across diverse underwater imaging scenarios.

In order to objectively assess the enhancement capability of the proposed AquaDiff model, we conduct quantitative evaluations across multiple benchmark underwater image datasets. These datasets U45, S16, and C60 represent diverse underwater conditions, ranging from mild color degradation to severe scattering and attenuation. To ensure a fair comparison, AquaDiff is evaluated alongside a broad selection of state-of-the-art UIE methods, including both traditional model-based techniques and recent deep learning–based approaches. The performance is measured using two widely adopted underwater image quality metrics, UIQM and UCIQE, which capture perceptual image quality, color fidelity, contrast, and structural integrity. The quantitative results summarized in Table~\ref{tab:quantitative_results} illustrate the superior consistency and robustness of AquaDiff across all datasets. A detailed analysis for each dataset is provided in this section below.

The U45 dataset represents a moderately challenging underwater imaging benchmark. The results show that AquaDiff achieves one of the strongest performances overall. In terms of UIQM, AquaDiff obtains a value of 4.609, which is highly competitive and slightly below only a few top-performing generative models such as UW-DDPM (4.899) and Water-Net (4.860). This indicates that AquaDiff preserves colorfulness, sharpness, and contrast at a level comparable to the best existing methods. However, AquaDiff achieves the highest UCIQE score of 0.5390 surpassing all methods including Ucolor (0.4461), DiffWater (0.4624), and UW-DDPM (0.4454). Because UCIQE is strongly influenced by color cast removal, chromatic balance, and edge contrast, this result reflects the ability of AquaDiff to correct wavelength-dependent distortions unique to underwater environments. Overall, despite a small gap in UIQM compared to the very top scores, AquaDiff delivers balanced and stable performance on U45 dataset.

\begin{figure*}[t!]
    \centering
    \includegraphics[width=\textwidth]{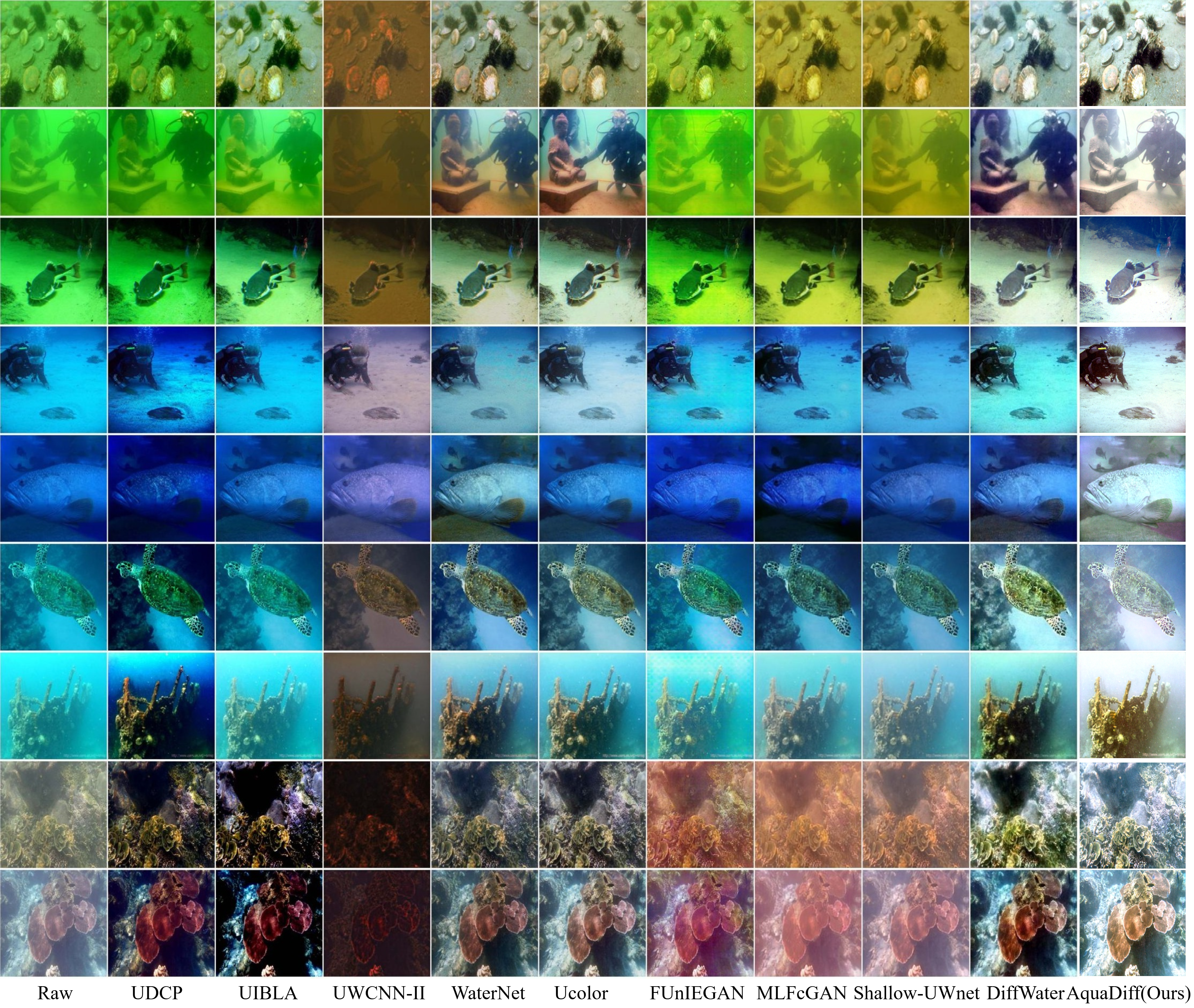}
    \caption{Qualitative comparison of our proposed method AquaDiff with traditional methods namely UDCP~\cite{drews2013transmission}, UIBLA~\cite{peng2017underwater} and deep learning-based methods namely UWCNN~\cite{li2020underwater}, Water-Net~\cite{li2019underwater}, Ucolor~\cite{li2021underwater}, MLFcGAN~\cite{liu2019mlfcgan}, FUnIEGAN~\cite{islam2020fast}, Shallow-uwnet~\cite{naik2021shallow}, DiffWater~\cite{guan2023diffwater} for underwater image enhancement on the U-45 dataset.}
    \label{fig:U45}
\end{figure*}

The S16 dataset is known for harsher degradation conditions, making it a rigorous test of enhancement robustness. AquaDiff attains a UIQM of 4.438 which is competitive but slightly lower than the DiffWater leading value of 4.5159. Nevertheless, AquaDiff achieves the highest UCIQE score on S16 at 0.5243 significantly outperforming all baselines including Ucolor (0.4514), UW-DDPM (0.4204), and MLcGAN (0.4172). This suggests that AquaDiff excels at enhancing chromatic contrast and luminance distribution even in scenes where scattering and attenuation are severe. The large improvement in UCIQE also indicates that AquaDiff yields visually more natural and color-consistent outputs compared to traditional and GAN-based approaches. Taken together, while DiffWater slightly surpasses AquaDiff in UIQM alone, AquaDiff exhibits the best overall enhancement quality due to its strong color correction and structural consistency. This reaffirms the advantages of its diffusion-driven generative modeling on challenging underwater conditions.

The C60 dataset contains some of the most complex underwater scenes, making performance differences between models more pronounced. AquaDiff achieves a UIQM of 4.324 placing it behind Water-Net (4.4453) and UW-DDPM, which produces the highest value of 4.8381. This indicates that some competing models may still outperform AquaDiff in sharpness preservation and global contrast under extremely challenging scenarios. However, AquaDiff once again demonstrates its core strength through chromatic enhancement, achieving the highest UCIQE score of 0.5176 outperforming DiffWater (0.4339), Ucolor (0.3846), and Water-Net (0.4416). Since C60 images typically suffer from severe color attenuation and wavelength imbalance. The superior UCIQE score of AquaDiff signals its effectiveness in recovering natural color distribution and reducing dominant color casts.

Despite not achieving the top UIQM, AquaDiff consistently provides the best color fidelity across all methods, illustrating that its design places strong emphasis on restoring perceptually meaningful color cues in real underwater environments.

The Figure~\ref{fig:AquaDiff_UIQM_UCIQE_radar_bars} presents a comparative evaluation of underwater image enhancement methods on the U45, S16, and C60 datasets using UIQM and UCIQE metrics. The radar chart on the left shows the performance of each method across all metrics, illustrating the strong and consistent results of AquaDiff (Ours). The bar charts on the right display the actual UIQM and UCIQE scores, where AquaDiff achieves high values across all datasets, demonstrating superior color correction, contrast improvement, and overall visual quality.

\subsection{Qualitative Evaluations}

While quantitative metrics provide objective measures of enhancement performance, qualitative visual assessment is essential for evaluating the perceptual quality and practical utility of UIE models. Visual inspection allows for the assessment of critical aspects such as color naturalness, detail preservation, artifact suppression, and overall visual appeal that may not be fully captured by numerical metrics alone. In this section, we present qualitative comparisons of AquaDiff against state-of-the-art methods across multiple test datasets, demonstrating the effectiveness of our model in addressing diverse underwater degradation scenarios including color cast correction, haze reduction, detail recovery, and natural appearance restoration. The visual comparisons encompass a wide variety of underwater scenes, including marine life, coral reefs, shipwrecks, diver activities, and man-made structures, captured under varying water clarity conditions, lighting scenarios, and depth levels.

To provide visual assessment of the enhancement capabilities across different underwater imaging scenarios, we present qualitative comparisons on the U90 test dataset. This test set encompasses diverse underwater scenes including marine life, coral reefs, shipwrecks, and diver activities under varying lighting and water conditions. Figure~\ref{fig:U90} illustrates representative results from eight distinct underwater scenes. Each scene demonstrating different degradation characteristics and enhancement challenges.

The qualitative analysis reveals significant improvements in color restoration and visual clarity achieved by the

\begin{figure*}[t!]
    \centering
    \includegraphics[width=\textwidth]{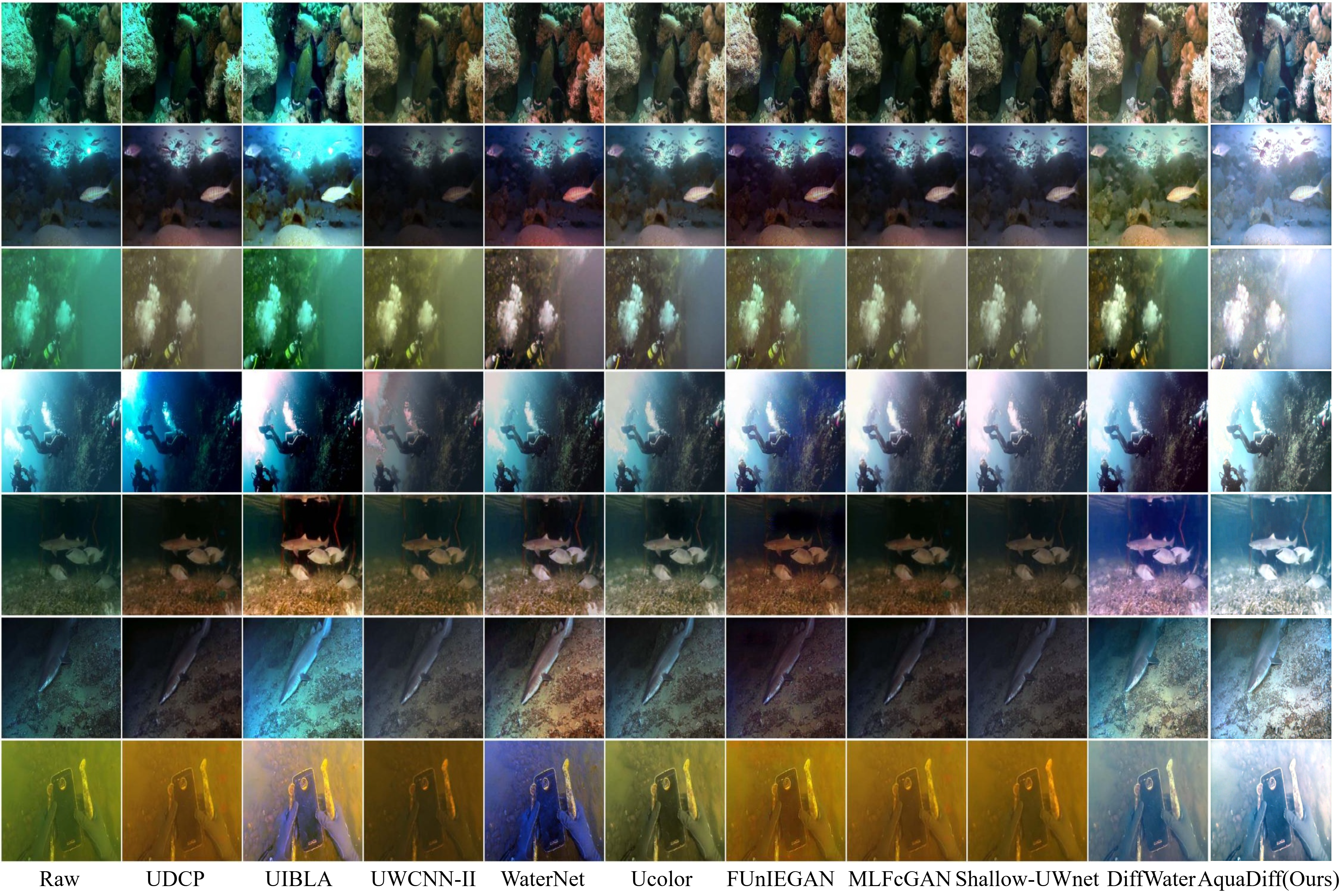}
    \caption{Qualitative comparison of our proposed method AquaDiff with traditional methods namely UDCP~\cite{drews2013transmission}, UIBLA~\cite{peng2017underwater} and deep learning-based methods namely UWCNN~\cite{li2020underwater}, Water-Net~\cite{li2019underwater}, Ucolor~\cite{li2021underwater}, MLFcGAN~\cite{liu2019mlfcgan}, FUnIEGAN~\cite{islam2020fast}, Shallow-uwnet~\cite{naik2021shallow}, DiffWater~\cite{guan2023diffwater} for underwater image enhancement on the C60 dataset.}
    \label{fig:C60}
\end{figure*}

proposed AquaDiff method compared to existing approaches. In scenes featuring close-up marine life and coral formations as seen in row 1 and row 4, AquaDiff effectively corrects severe color casts ranging from dominant blue-green tints to reddish-brown distortions. Furthermore, our model AquaDiff restores natural color tones while preserving fine textures and intricate details of the subjects. Our model demonstrates particular strength in handling scenes with artificial lighting, such as the shark resting on the seabed (Row 8), where it successfully recovers shadow details and restores natural coloration that remains obscured or monochromatic in outputs from competing methods.

Haze reduction and visibility improvement represent another key advantage of AquaDiff as evidenced in wide-angle reef scenes as seen in row 5 and row 7 and shipwreck imagery as seen in row 2. While traditional methods such as UDCP and UIBLA often introduce artifacts or fail to adequately address atmospheric scattering. On the other hand, the CNN-based approaches such as UWCNN and Ucolor tend to produce over-enhanced or under-corrected results, AquaDiff achieves a balanced enhancement that removes haze while maintaining natural appearance and preventing color oversaturation. Our model has the ability to handle varying water clarity conditions is particularly notable in diver scenes as seen in row 3 and row 6. In these scenes, our model preserves the natural blue-green water tones while significantly improving visibility of distant objects and fine details such as equipment textures and sand grain patterns.

Furthermore, AquaDiff demonstrates superior performance in scenes with complex lighting conditions, such as natural light filtering from the surface as seen in row 5 and mixed artificial-natural illumination as seen in row 8. The diffusion-based approach effectively handles the wide dynamic range present in these scenarios. This leads to recovering details in both shadow and highlight regions without introducing halos or unnatural brightness transitions that are commonly observed in GAN-based methods such as MLFcGAN and FUnIEGAN. The preservation of light rays and natural illumination patterns, combined with improved contrast and color fidelity. This results in visually pleasing outputs that closely approximate the expected appearance of underwater scenes under optimal imaging conditions.

Qualitative evaluation on the U45 test dataset, as visually represented in Figure~\ref{fig:U45}, further substantiates AquaDiff's robust enhancement capabilities across diverse underwater scenarios. 

The method consistently excels in correcting severe color casts, effectively transforming images dominated by green-yellow hues for sandy seabeds, diver scenes in rows 1-3. Moreover, or model removes the deep blue saturations for marine life and divers in rows 4-6. Our model removes the reddish-brown distortions for coral reefs in row 8 into visually natural representations. Beyond color restoration, AquaDiff significantly reduces haze and improves overall clarity. Thereby, enhancing visibility in complex scenes such as shipwrecks in row 7 and intricate coral formations. This leads to superior detail preservation. This results in images that reveal fine textures on marine life and environmental structures that were previously obscured. The enhanced images consistently exhibit natural color balance, improved contrast, and sharper details.

\begin{figure*}[t!]
    \centering
    \includegraphics[width=\textwidth]{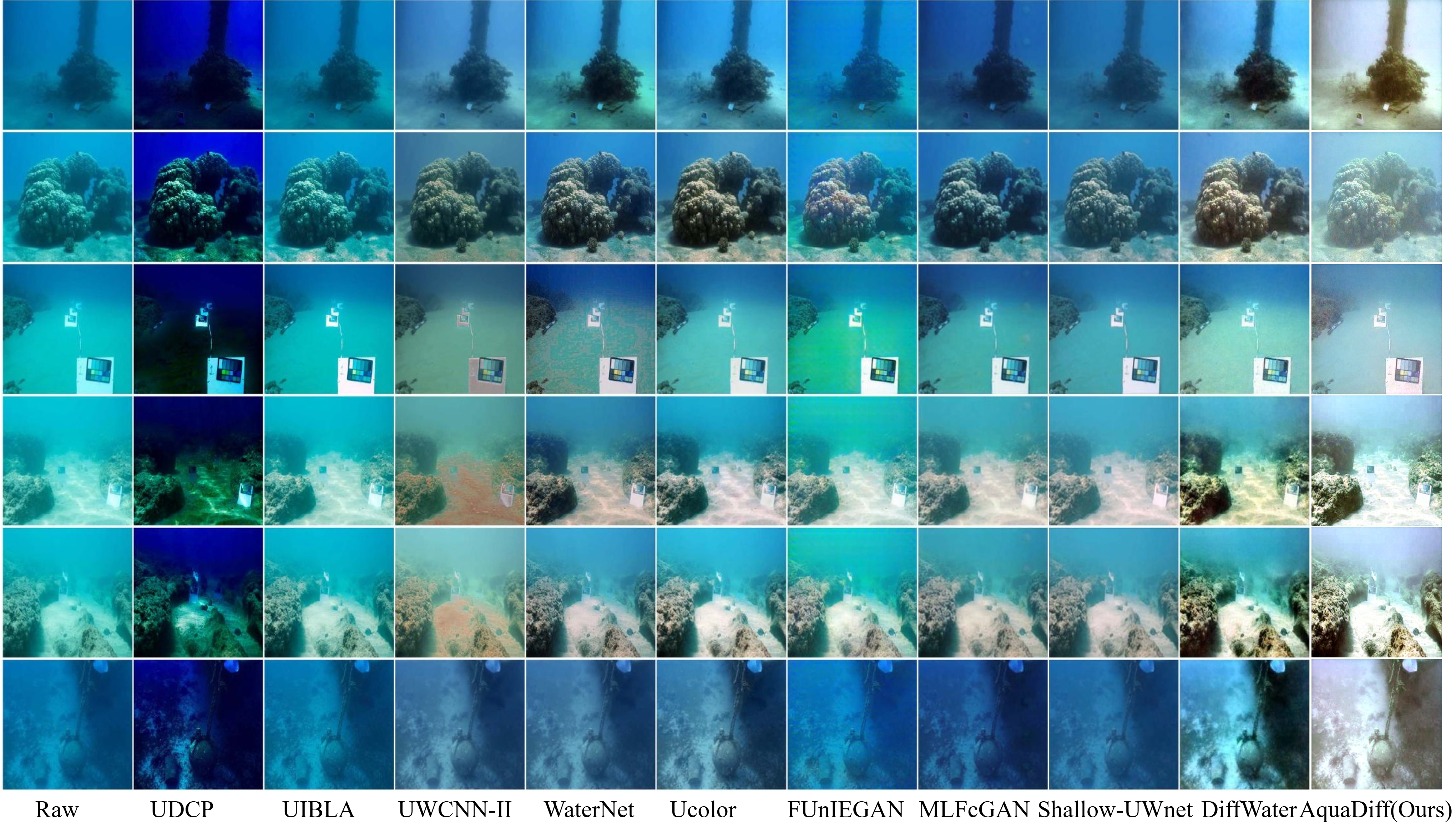}
    \caption{Qualitative comparison of our proposed method AquaDiff with traditional methods namely UDCP~\cite{drews2013transmission}, UIBLA~\cite{peng2017underwater} and deep learning-based methods namely UWCNN~\cite{li2020underwater}, Water-Net~\cite{li2019underwater}, Ucolor~\cite{li2021underwater}, MLFcGAN~\cite{liu2019mlfcgan}, FUnIEGAN~\cite{islam2020fast}, Shallow-uwnet~\cite{naik2021shallow}, DiffWater~\cite{guan2023diffwater} for underwater image enhancement on the S-16 dataset.}
    \label{fig:S16}
\end{figure*}

Qualitative evaluation on the S16 test dataset, as illustrated in Figure~\ref{fig:S16} demonstrates the effectiveness of AquaDiff in handling diverse and challenging underwater degradation scenarios. Our model successfully addresses severe color distortions across multiple scene types including the correction of deep blue-green casts in pier structures and coral formations as seen in row 1 and row 2, the removal of intense green turbidity in diver scenes with color calibration charts as shown in row 3. The neutralization of reddish-brown distortions in highly turbid water conditions in row 4. Particularly noteworthy is the performance of AquaDiff in scenes with color calibration charts where the method accurately restores distinct color squares namely reds, blues, greens, yellows that are completely obscured in the original degraded images. This demonstrates precise color correction capabilities. Our model achieves substantial haze reduction across all scenes, significantly improving visibility of structural details such as pier foundations, coral textures, rock formations, and metallic objects like anchors and chains. This enhanced clarity enables the recovery of fine details including seabed textures, diver equipment, and intricate coral patterns that remain indistinct in outputs from competing methods.

The qualitative evaluation on the C60 test dataset, as illustrated in Figure~\ref{fig:C60} provides  visual comparison of enhancement outputs across different methods, revealing distinct performance characteristics and quality differences. 

Through detailed examination of representative scenes, we observe significant variations in how each method addresses the multifaceted challenges of underwater image degradation.

In blue-green dominant scenes, traditional methods produce unnatural or desaturated outputs that fail to correct strong casts. CNN- and GAN-based approaches show improvements but often introduce color imbalances, artifacts, or inconsistent corrections across regions. In contrast, AquaDiff consistently restores accurate and natural colors while effectively removing blue-green casts. In green-yellow turbid scenes, traditional methods struggle with incomplete correction and unnatural shifts, leaving strong residual tints. CNN- and GAN-based methods improve performance but may introduce artifacts, over-enhancement, or inconsistent correction across the scene. AquaDiff reliably neutralizes green-yellow casts while preserving natural sandy tones and realistic color variations In reddish-brown turbid environments, traditional and CNN-based methods fail to fully correct distortions or maintain natural color balance. GAN-based methods reduce turbidity but may introduce artifacts or inconsistencies across textured regions. AquaDiff consistently removes reddish-brown casts and restores natural colors without oversaturation. 

For haze reduction, traditional methods provide minimal improvement and often introduce halos around high-contrast edges. CNN- and GAN-based approaches improve visibility but may over-enhance brightness or leave residual haze in dense regions. AquaDiff achieves superior haze removal with natural appearance, revealing distant objects and fine structural details. Regarding detail preservation, traditional methods blur textures and lose important features in corals, vegetation, and seabed structures. CNN- and GAN-based methods show better recovery but may introduce blurring, oversharpening, or checkerboard artifacts. AquaDiff maintains natural sharpness and recovers both global structures and fine textures effectively. Artifact analysis shows that traditional methods introduce halos and banding, and CNN/GAN methods may create over-enhancement or unrealistic textures. Water-Net performs better but still struggles under extreme conditions. AquaDiff produces clean, artifact-free outputs with natural contrast, sharp edges, and coherent brightness distribution.

\subsection{Ablation Studies}

To analyze the contribution of individual components in the proposed AquaDiff framework, we conduct ablation studies focusing on the effectiveness of the cross-domain consistency loss and the impact of the enhanced U-Net denoising backbone. All ablation variants are trained and evaluated under identical settings using the same datasets, training schedules, and evaluation metrics to ensure a fair comparison. Performance is reported using the widely adopted no-reference underwater image quality metrics UIQM and UCIQE, which assess perceptual quality, color fidelity, and structural integrity.

\subsubsection{Effectiveness of Cross-Domain Consistency Loss}

The cross-domain consistency loss (CDCL) is designed to jointly enforce pixel-level fidelity, perceptual similarity, structural consistency, and frequency-domain coherence during diffusion-based enhancement. To evaluate its effectiveness, we compare the full AquaDiff model against a variant trained without CDCL, where only a standard pixel-wise $\ell_1$ loss is employed. As shown in Table~\ref{tab:ablation_combined}, removing CDCL results in noticeable degradation in both UIQM and UCIQE scores across all evaluated datasets. In particular, the reduction in UCIQE indicates inferior color correction and contrast restoration, while the drop in UIQM reflects weakened sharpness and structural preservation. These results demonstrate that CDCL plays a critical role in stabilizing diffusion training and suppressing common artifacts, leading to visually coherent and perceptually faithful underwater image enhancement.

\subsubsection{Contribution of Enhanced UNet backbone}

To assess the contribution of the enhanced U-Net backbone, we replace it with a standard diffusion U-Net architecture that excludes residual dense blocks, dense skip connections, and multi-resolution attention modules. Quantitative results in Table~\ref{tab:ablation_combined} show that the enhanced backbone consistently improves performance. The inclusion of residual dense blocks strengthens feature reuse and gradient flow, while multi-resolution attention improves long-range dependency modeling and global color consistency. These architectural enhancements enable the model to better capture complex underwater degradation patterns, resulting in improved color fidelity, sharper details, and more stable enhancement compared to the baseline U-Net design.

\begin{table}[t]
\centering
\caption{Combined ablation study evaluating the impact of the cross-domain consistency loss and the enhanced U-Net backbone.}
\label{tab:ablation_combined}
\renewcommand{\arraystretch}{1.15}
\setlength{\tabcolsep}{6pt}
\begin{tabular}{lcc}
\toprule
\textbf{Model Variant} &
\textbf{UIQM} $\uparrow$ &
\textbf{UCIQE} $\uparrow$ \\
\midrule
Baseline Diffusion Model      & 4.12 & 0.486 \\
+ CDCL Only                   & 4.38 & 0.521 \\
+ Enhanced U-Net Only         & 4.45 & 0.528 \\
AquaDiff (Full Model)         & \textbf{4.61} & \textbf{0.539} \\
\bottomrule
\end{tabular}
\end{table}

\section{Conclusions}
\label{sec:conclusion}

This paper presented AquaDiff, a diffusion-based underwater image enhancement framework designed to address wavelength-dependent color distortion while preserving structural and perceptual fidelity. By integrating chromatic prior-guided color compensation with a conditional diffusion process, AquaDiff effectively leverages cross-attention to fuse degraded inputs and noisy latent representations across diffusion timesteps. The enhanced denoising backbone, combined with residual dense blocks and multi-resolution attention, enables robust recovery of both global color consistency and fine-grained details. Furthermore, the proposed cross-domain consistency loss jointly constrains pixel, perceptual, structural, and frequency domains, mitigating common diffusion artifacts and improving visual coherence.

Extensive quantitative and qualitative evaluations on multiple challenging underwater benchmarks demonstrate that AquaDiff consistently achieves superior color fidelity and competitive overall image quality compared to state-of-the-art traditional, CNN-, GAN-, and diffusion-based methods. These results highlight the effectiveness of diffusion models for underwater image enhancement and establish AquaDiff as a robust and generalizable solution for real-world underwater vision applications.

\bibliographystyle{IEEEbib}
\bibliography{strings,ref}

\begin{thebibliography}{10}

\bibitem{saoud2024seeing}
Lyes~Saad Saoud, Mahmoud Elmezain, Atif Sultan, Mohamed Heshmat, Lakmal Seneviratne, and Irfan Hussain,
\newblock ``Seeing through the haze: A comprehensive review of underwater image enhancement techniques,''
\newblock {\em IEEE Access}, 2024.

\bibitem{qi2021underwater}
Qi~Qi, Yongchang Zhang, Fei Tian, QM~Jonathan Wu, Kunqian Li, Xin Luan, and Dalei Song,
\newblock ``Underwater image co-enhancement with correlation feature matching and joint learning,''
\newblock {\em IEEE Transactions on Circuits and Systems for Video Technology}, vol. 32, no. 3, pp. 1133--1147, 2021.

\bibitem{sun2023high}
Sibo Sun, Huigong Guo, Guangming Wan, Chao Dong, Ce~Zheng, and Yong Wang,
\newblock ``High-precision underwater acoustic localization of the black box utilizing an autonomous underwater vehicle based on the improved artificial potential field,''
\newblock {\em IEEE Transactions on Geoscience and Remote Sensing}, vol. 61, pp. 1--10, 2023.

\bibitem{hao2024underwater}
Yansheng Hao, Yaoyao Yuan, Hongman Zhang, and Ze~Zhang,
\newblock ``Underwater optical imaging: methods, applications and perspectives,''
\newblock {\em Remote Sensing}, vol. 16, no. 20, pp. 3773, 2024.

\bibitem{galetto2025deep}
Fernando Galetto and Guang Deng,
\newblock ``A deep learning approach for marine snow synthesis and removal,''
\newblock {\em Signal, Image and Video Processing}, vol. 19, no. 1, pp. 1, 2025.

\bibitem{alsakar2025underwater}
Yasmin~M Alsakar, Nehal~A Sakr, Shaker El-Sappagh, Tamer Abuhmed, and Mohammed Elmogy,
\newblock ``Underwater image restoration and enhancement: a comprehensive review of recent trends, challenges, and applications,''
\newblock {\em The Visual Computer}, vol. 41, no. 6, pp. 3735--3783, 2025.

\bibitem{zhou2022yolotrashcan}
Wei Zhou, Fujian Zheng, Gang Yin, Yiran Pang, and Jun Yi,
\newblock ``Yolotrashcan: A deep learning marine debris detection network,''
\newblock {\em IEEE Transactions on Instrumentation and Measurement}, vol. 72, pp. 1--12, 2022.

\bibitem{cao2022dynamic}
Xiang Cao, Lu~Ren, and Changyin Sun,
\newblock ``Dynamic target tracking control of autonomous underwater vehicle based on trajectory prediction,''
\newblock {\em IEEE Transactions on Cybernetics}, vol. 53, no. 3, pp. 1968--1981, 2022.

\bibitem{naveen2025advancements}
Palanichamy Naveen,
\newblock ``Advancements in underwater imaging through machine learning: Techniques, challenges, and applications,''
\newblock {\em Multimedia Tools and Applications}, , no. 22, pp. 24839--24858, 2025.

\bibitem{almutiry2024underwater}
Omar Almutiry, Khalid Iqbal, Shariq Hussain, Awais Mahmood, and Habib Dhahri,
\newblock ``Underwater images contrast enhancement and its challenges: a survey,''
\newblock {\em Multimedia Tools and Applications}, vol. 83, no. 5, pp. 15125--15150, 2024.

\bibitem{zhou2025uw}
Wenhao Zhou, Junbao Zeng, Shuo Li, and Yuexing Zhang,
\newblock ``Uw-yolo-bio: A real-time lightweight detector for underwater biological perception with global and regional context awareness,''
\newblock {\em Journal of Marine Science and Engineering}, vol. 13, no. 11, pp. 2189, 2025.

\bibitem{huang2025visual}
Guoxi Huang, Haoran Wang, Brett Seymour, Evan Kovacs, John Ellerbrock, Dave Blackham, and Nantheera Anantrasirichai,
\newblock ``Visual enhancement and 3d representation for underwater scenes: a review,''
\newblock {\em arXiv e-prints}, pp. arXiv--2505, 2025.

\bibitem{summers2025impact}
Jason~M Summers, Mark~W Jones, and Catherine Seale,
\newblock ``Impact of underwater image enhancement on feature matching,''
\newblock {\em Sensors}, vol. 25, no. 22, pp. 6966, 2025.

\bibitem{zhang2023improved}
Jian Zhang, Jinshuai Zhang, Kexin Zhou, Yonghui Zhang, Hongda Chen, and Xinyue Yan,
\newblock ``An improved yolov5-based underwater object-detection framework,''
\newblock {\em Sensors}, vol. 23, no. 7, pp. 3693, 2023.

\bibitem{jyothimurugan2025efficient}
Mohan Jyothimurugan, S~Pavithra, and J~Deepika~Roselind,
\newblock ``Efficient underwater ecological monitoring with embedded ai: detecting crown-of-thorns starfish via dcgan and yolov6,''
\newblock {\em Frontiers in Marine Science}, vol. 12, pp. 1658205, 2025.

\bibitem{hummel1975image}
Robert Hummel,
\newblock ``Image enhancement by histogram transformation,''
\newblock {\em Unknown}, 1975.

\bibitem{ancuti2017color}
Codruta~O Ancuti, Cosmin Ancuti, Christophe De~Vleeschouwer, and Philippe Bekaert,
\newblock ``Color balance and fusion for underwater image enhancement,''
\newblock {\em IEEE Transactions on image processing}, vol. 27, no. 1, pp. 379--393, 2017.

\bibitem{jin2001contrast}
Yinpeng Jin, Laura~M Fayad, and Andrew~F Laine,
\newblock ``Contrast enhancement by multiscale adaptive histogram equalization,''
\newblock in {\em Wavelets: Applications in Signal and Image Processing IX}. SPIE, 2001, vol. 4478, pp. 206--213.

\bibitem{peng2018multi}
LI~Peng, Yong HUANG, and YAO Kunlun,
\newblock ``Multi-algorithm fusion of rgb and hsv color spaces for image enhancement,''
\newblock in {\em 2018 37th Chinese Control Conference (CCC)}. IEEE, 2018, pp. 9584--9589.

\bibitem{drews2013transmission}
Paul Drews, Erickson Nascimento, Filipe Moraes, Silvia Botelho, and Mario Campos,
\newblock ``Transmission estimation in underwater single images,''
\newblock in {\em Proceedings of the IEEE international conference on computer vision workshops}, 2013, pp. 825--830.

\bibitem{berman2020underwater}
Dana Berman, Deborah Levy, Shai Avidan, and Tali Treibitz,
\newblock ``Underwater single image color restoration using haze-lines and a new quantitative dataset,''
\newblock {\em IEEE transactions on pattern analysis and machine intelligence}, vol. 43, no. 8, pp. 2822--2837, 2020.

\bibitem{peng2017underwater}
Yan-Tsung Peng and Pamela~C Cosman,
\newblock ``Underwater image restoration based on image blurriness and light absorption,''
\newblock {\em IEEE transactions on image processing}, vol. 26, no. 4, pp. 1579--1594, 2017.

\bibitem{yuan2020underwater}
Jieyu Yuan, Wei Cao, Zhanchuan Cai, and Binghua Su,
\newblock ``An underwater image vision enhancement algorithm based on contour bougie morphology,''
\newblock {\em IEEE Transactions on Geoscience and Remote Sensing}, vol. 59, no. 10, pp. 8117--8128, 2020.

\bibitem{ren2022reinforced}
Tingdi Ren, Haiyong Xu, Gangyi Jiang, Mei Yu, Xuan Zhang, Biao Wang, and Ting Luo,
\newblock ``Reinforced swin-convs transformer for simultaneous underwater sensing scene image enhancement and super-resolution,''
\newblock {\em IEEE Transactions on Geoscience and Remote Sensing}, vol. 60, pp. 1--16, 2022.

\bibitem{shen2023udaformer}
Zhen Shen, Haiyong Xu, Ting Luo, Yang Song, and Zhouyan He,
\newblock ``Udaformer: Underwater image enhancement based on dual attention transformer,''
\newblock {\em Computers \& Graphics}, vol. 111, pp. 77--88, 2023.

\bibitem{goodfellow2020generative}
Ian Goodfellow, Jean Pouget-Abadie, Mehdi Mirza, Bing Xu, David Warde-Farley, Sherjil Ozair, Aaron Courville, and Yoshua Bengio,
\newblock ``Generative adversarial networks,''
\newblock {\em Communications of the ACM}, vol. 63, no. 11, pp. 139--144, 2020.

\bibitem{fu2022uncertainty}
Zhenqi Fu, Wu~Wang, Yue Huang, Xinghao Ding, and Kai-Kuang Ma,
\newblock ``Uncertainty inspired underwater image enhancement,''
\newblock in {\em European conference on computer vision}. Springer, 2022, pp. 465--482.

\bibitem{li2020underwater}
Chongyi Li, Saeed Anwar, and Fatih Porikli,
\newblock ``Underwater scene prior inspired deep underwater image and video enhancement,''
\newblock {\em Pattern recognition}, vol. 98, pp. 107038, 2020.

\bibitem{li2019underwater}
Chongyi Li, Chunle Guo, Wenqi Ren, Runmin Cong, Junhui Hou, Sam Kwong, and Dacheng Tao,
\newblock ``An underwater image enhancement benchmark dataset and beyond,''
\newblock {\em IEEE transactions on image processing}, vol. 29, pp. 4376--4389, 2019.

\bibitem{li2021underwater}
Chongyi Li, Saeed Anwar, Junhui Hou, Runmin Cong, Chunle Guo, and Wenqi Ren,
\newblock ``Underwater image enhancement via medium transmission-guided multi-color space embedding,''
\newblock {\em IEEE Transactions on Image Processing}, vol. 30, pp. 4985--5000, 2021.

\bibitem{liu2019mlfcgan}
Xiaodong Liu, Zhi Gao, and Ben~M Chen,
\newblock ``Mlfcgan: Multilevel feature fusion-based conditional gan for underwater image color correction,''
\newblock {\em IEEE Geoscience and Remote Sensing Letters}, vol. 17, no. 9, pp. 1488--1492, 2019.

\bibitem{jahidul2019fast}
Md~Jahidul~Islam, Youya Xia, and Junaed Sattar,
\newblock ``Fast underwater image enhancement for improved visual perception,''
\newblock {\em arXiv e-prints}, pp. arXiv--1903, 2019.

\bibitem{naik2021shallow}
Ankita Naik, Apurva Swarnakar, and Kartik Mittal,
\newblock ``Shallow-uwnet: Compressed model for underwater image enhancement (student abstract),''
\newblock in {\em Proceedings of the AAAI Conference on Artificial Intelligence}, 2021, vol.~35, pp. 15853--15854.

\bibitem{han2023uiegan}
Guangjie Han, Min Wang, Hongbo Zhu, and Chuan Lin,
\newblock ``Uiegan: Adversarial learning-based photorealistic image enhancement for intelligent underwater environment perception,''
\newblock {\em IEEE Transactions on Geoscience and Remote Sensing}, vol. 61, pp. 1--14, 2023.

\bibitem{yuan2021tebcf}
Jieyu Yuan, Zhanchuan Cai, and Wei Cao,
\newblock ``Tebcf: Real-world underwater image texture enhancement model based on blurriness and color fusion,''
\newblock {\em IEEE Transactions on Geoscience and Remote Sensing}, vol. 60, pp. 1--15, 2021.

\bibitem{chen2021mffn}
Renzhang Chen, Zhanchuan Cai, and Wei Cao,
\newblock ``Mffn: An underwater sensing scene image enhancement method based on multiscale feature fusion network,''
\newblock {\em IEEE Transactions on Geoscience and Remote Sensing}, vol. 60, pp. 1--12, 2021.

\bibitem{jiang2022two}
Qun Jiang, Yunfeng Zhang, Fangxun Bao, Xiuyang Zhao, Caiming Zhang, and Peide Liu,
\newblock ``Two-step domain adaptation for underwater image enhancement,''
\newblock {\em Pattern Recognition}, vol. 122, pp. 108324, 2022.

\bibitem{fabbri2018enhancing}
Cameron Fabbri, Md~Jahidul Islam, and Junaed Sattar,
\newblock ``Enhancing underwater imagery using generative adversarial networks,''
\newblock in {\em 2018 IEEE international conference on robotics and automation (ICRA)}. IEEE, 2018, pp. 7159--7165.

\bibitem{liang2021gudcp}
Zheng Liang, Xueyan Ding, Yafei Wang, Xiaohong Yan, and Xianping Fu,
\newblock ``Gudcp: Generalization of underwater dark channel prior for underwater image restoration,''
\newblock {\em IEEE transactions on circuits and systems for video technology}, vol. 32, no. 7, pp. 4879--4884, 2021.

\bibitem{fu2014retinex}
Xueyang Fu, Peixian Zhuang, Yue Huang, Yinghao Liao, Xiao-Ping Zhang, and Xinghao Ding,
\newblock ``A retinex-based enhancing approach for single underwater image,''
\newblock in {\em 2014 IEEE international conference on image processing (ICIP)}. Ieee, 2014, pp. 4572--4576.

\bibitem{ancuti2012enhancing}
Cosmin Ancuti, Codruta~Orniana Ancuti, Tom Haber, and Philippe Bekaert,
\newblock ``Enhancing underwater images and videos by fusion,''
\newblock in {\em 2012 IEEE conference on computer vision and pattern recognition}. IEEE, 2012, pp. 81--88.

\bibitem{gao2021underwater}
Farong Gao, Kai Wang, Zhangyi Yang, Yejian Wang, and Qizhong Zhang,
\newblock ``Underwater image enhancement based on local contrast correction and multi-scale fusion,''
\newblock {\em Journal of Marine Science and Engineering}, vol. 9, no. 2, pp. 225, 2021.

\bibitem{zhang2024underwater}
Weidong Zhang, Qingmin Liu, Yikun Feng, Lei Cai, and Peixian Zhuang,
\newblock ``Underwater image enhancement via principal component fusion of foreground and background,''
\newblock {\em IEEE Transactions on Circuits and Systems for Video Technology}, vol. 34, no. 11, pp. 10930--10943, 2024.

\bibitem{xu2016single}
Yueshu Xu, Xiaoqiang Guo, Haiying Wang, Fang Zhao, and Longfei Peng,
\newblock ``Single image haze removal using light and dark channel prior,''
\newblock in {\em 2016 IEEE/CIC International Conference on Communications in China (ICCC)}. IEEE, 2016, pp. 1--6.

\bibitem{peng2018generalization}
Yan-Tsung Peng, Keming Cao, and Pamela~C Cosman,
\newblock ``Generalization of the dark channel prior for single image restoration,''
\newblock {\em IEEE Transactions on Image Processing}, vol. 27, no. 6, pp. 2856--2868, 2018.

\bibitem{lu2015contrast}
Huimin Lu, Yujie Li, Lifeng Zhang, and Seiichi Serikawa,
\newblock ``Contrast enhancement for images in turbid water,''
\newblock {\em Journal of the Optical Society of America A}, vol. 32, no. 5, pp. 886--893, 2015.

\bibitem{zhao2015deriving}
Xinwei Zhao, Tao Jin, and Song Qu,
\newblock ``Deriving inherent optical properties from background color and underwater image enhancement,''
\newblock {\em Ocean Engineering}, vol. 94, pp. 163--172, 2015.

\bibitem{akkaynak2018revised}
Derya Akkaynak and Tali Treibitz,
\newblock ``A revised underwater image formation model,''
\newblock in {\em Proceedings of the IEEE conference on computer vision and pattern recognition}, 2018, pp. 6723--6732.

\bibitem{mcglamery1980computer}
BL~McGlamery,
\newblock ``A computer model for underwater camera systems,''
\newblock in {\em Ocean optics VI}. SPIE, 1980, vol. 208, pp. 221--231.

\bibitem{jaffe2002computer}
Jules~S Jaffe,
\newblock ``Computer modeling and the design of optimal underwater imaging systems,''
\newblock {\em IEEE journal of oceanic engineering}, vol. 15, no. 2, pp. 101--111, 2002.

\bibitem{akkaynak2019sea}
Derya Akkaynak and Tali Treibitz,
\newblock ``Sea-thru: A method for removing water from underwater images,''
\newblock in {\em Proceedings of the IEEE/CVF conference on computer vision and pattern recognition}, 2019, pp. 1682--1691.

\bibitem{li2024taformer}
Yuanyuan Li, Zetian Mi, Yulin Wang, Shuaiyong Jiang, and Xianping Fu,
\newblock ``Taformer: A transmission-aware transformer for underwater image enhancement,''
\newblock {\em IEEE Transactions on Circuits and Systems for Video Technology}, 2024.

\bibitem{zhang2022underwater}
Weidong Zhang, Yudong Wang, and Chongyi Li,
\newblock ``Underwater image enhancement by attenuated color channel correction and detail preserved contrast enhancement,''
\newblock {\em IEEE Journal of Oceanic Engineering}, vol. 47, no. 3, pp. 718--735, 2022.

\bibitem{cong2023pugan}
Runmin Cong, Wenyu Yang, Wei Zhang, Chongyi Li, Chun-Le Guo, Qingming Huang, and Sam Kwong,
\newblock ``Pugan: Physical model-guided underwater image enhancement using gan with dual-discriminators,''
\newblock {\em IEEE Transactions on Image Processing}, vol. 32, pp. 4472--4485, 2023.

\bibitem{ho2020denoising}
Jonathan Ho, Ajay Jain, and Pieter Abbeel,
\newblock ``Denoising diffusion probabilistic models,''
\newblock {\em Advances in neural information processing systems}, vol. 33, pp. 6840--6851, 2020.

\bibitem{nichol2021improved}
Alexander~Quinn Nichol and Prafulla Dhariwal,
\newblock ``Improved denoising diffusion probabilistic models,''
\newblock in {\em International conference on machine learning}. PMLR, 2021, pp. 8162--8171.

\bibitem{ho2022cascaded}
Jonathan Ho, Chitwan Saharia, William Chan, David~J Fleet, Mohammad Norouzi, and Tim Salimans,
\newblock ``Cascaded diffusion models for high fidelity image generation,''
\newblock {\em Journal of Machine Learning Research}, vol. 23, no. 47, pp. 1--33, 2022.

\bibitem{rombach2022high}
Robin Rombach, Andreas Blattmann, Dominik Lorenz, Patrick Esser, and Bj{\"o}rn Ommer,
\newblock ``High-resolution image synthesis with latent diffusion models,''
\newblock in {\em Proceedings of the IEEE/CVF conference on computer vision and pattern recognition}, 2022, pp. 10684--10695.

\bibitem{mardani2023variational}
Morteza Mardani, Jiaming Song, Jan Kautz, and Arash Vahdat,
\newblock ``A variational perspective on solving inverse problems with diffusion models,''
\newblock {\em arXiv preprint arXiv:2305.04391}, 2023.

\bibitem{ozdenizci2023restoring}
Ozan Ozdenizci and Robert Legenstein,
\newblock ``Restoring vision in adverse weather conditions with patch-based denoising diffusion models,''
\newblock {\em IEEE Transactions on Pattern Analysis \& Machine Intelligence}, vol. 45, no. 08, pp. 10346--10357, 2023.

\bibitem{kawar2022denoising}
Bahjat Kawar, Michael Elad, Stefano Ermon, and Jiaming Song,
\newblock ``Denoising diffusion restoration models,''
\newblock {\em Advances in neural information processing systems}, vol. 35, pp. 23593--23606, 2022.

\bibitem{sohl2015deep}
Jascha Sohl-Dickstein, Eric Weiss, Niru Maheswaranathan, and Surya Ganguli,
\newblock ``Deep unsupervised learning using nonequilibrium thermodynamics,''
\newblock in {\em International conference on machine learning}. pmlr, 2015, pp. 2256--2265.

\bibitem{saharia2022image}
Chitwan Saharia, Jonathan Ho, William Chan, Tim Salimans, David~J Fleet, and Mohammad Norouzi,
\newblock ``Image super-resolution via iterative refinement,''
\newblock {\em IEEE transactions on pattern analysis and machine intelligence}, vol. 45, no. 4, pp. 4713--4726, 2022.

\bibitem{lu2023speed}
Siqi Lu, Fengxu Guan, Hanyu Zhang, and Haitao Lai,
\newblock ``Speed-up ddpm for real-time underwater image enhancement,''
\newblock {\em IEEE Transactions on Circuits and Systems for Video Technology}, vol. 34, no. 5, pp. 3576--3588, 2023.

\bibitem{guan2023diffwater}
Meisheng Guan, Haiyong Xu, Gangyi Jiang, Mei Yu, Yeyao Chen, Ting Luo, and Xuebo Zhang,
\newblock ``Diffwater: Underwater image enhancement based on conditional denoising diffusion probabilistic model,''
\newblock {\em IEEE Journal of Selected Topics in Applied Earth Observations and Remote Sensing}, vol. 17, pp. 2319--2335, 2023.

\bibitem{zhang2023metaue}
Zhenwei Zhang, Haorui Yan, Ke~Tang, and Yuping Duan,
\newblock ``Metaue: Model-based meta-learning for underwater image enhancement,''
\newblock {\em arXiv preprint arXiv:2303.06543}, 2023.

\bibitem{ancuti2019color}
Codruta~O Ancuti, Cosmin Ancuti, Christophe De~Vleeschouwer, and Mateu Sbert,
\newblock ``Color channel compensation (3c): A fundamental pre-processing step for image enhancement,''
\newblock {\em IEEE Transactions on Image Processing}, vol. 29, pp. 2653--2665, 2019.

\bibitem{peng2023u}
Lintao Peng, Chunli Zhu, and Liheng Bian,
\newblock ``U-shape transformer for underwater image enhancement,''
\newblock {\em IEEE transactions on image processing}, vol. 32, pp. 3066--3079, 2023.

\bibitem{islam2020fast}
Md~Jahidul Islam, Youya Xia, and Junaed Sattar,
\newblock ``Fast underwater image enhancement for improved visual perception,''
\newblock {\em IEEE robotics and automation letters}, vol. 5, no. 2, pp. 3227--3234, 2020.

\bibitem{li2019fusion}
Hanyu Li, Jingjing Li, and Wei Wang,
\newblock ``A fusion adversarial underwater image enhancement network with a public test dataset,''
\newblock {\em arXiv preprint arXiv:1906.06819}, 2019.

\bibitem{berman2021squid}
Dana Berman, Deborah Levy, Shai Avidan, and Tali Treibitz,
\newblock ``Squid-stereo quantitative underwater image dataset,''
\newblock {\em Zenodo https://doi. org/10.5281/zenodo}, vol. 5744037, 2021.

\bibitem{a65}
Zhou Wang, A.C. Bovik, H.R. Sheikh, and E.P. Simoncelli,
\newblock ``Image quality assessment: from error visibility to structural similarity,''
\newblock {\em IEEE Transactions on Image Processing}, vol. 13, no. 4, pp. 600--612, 2004.

\bibitem{a66}
Karen Panetta, Chen Gao, and Sos Agaian,
\newblock ``Human-visual-system-inspired underwater image quality measures,''
\newblock {\em IEEE Journal of Oceanic Engineering}, vol. 41, no. 3, pp. 541--551, 2016.

\end{thebibliography}

\end{document}